\documentclass[runningheads]{llncs}
\usepackage[utf8]{inputenc}
\usepackage{graphicx}
\usepackage[misc]{ifsym}
\usepackage{url}
\usepackage[english]{babel}
\usepackage{bbding}
\usepackage{rotating}
\usepackage{verbatim}
\usepackage{multirow}
\usepackage{algorithm} 
\usepackage{amsmath}
\usepackage{algpseudocode}

\usepackage{amsthm}
\theoremstyle{definition}
\usepackage[flushleft]{threeparttable}
\usepackage{colortbl}
\usepackage[table]{xcolor}
\usepackage{lipsum}
\usepackage{amssymb}
\usepackage{diagbox}
\usepackage{subfigure}
\usepackage{hyperref}
\algnewcommand\algorithmicforeach{\textbf{for each}}
\algdef{S}[FOR]{ForEach}[1]{\algorithmicforeach\ #1\ \algorithmicdo}
\newtheorem{mydef}{Definition}
\newcommand{\el}{{\cal {EL}}}

\newcommand{\elb}{{\mathcal{EL_\perp}}}

\begin{document}

\title{Repairing Networks of $\mathcal{EL_\perp}$ Ontologies using Weakening and Completing \\ - \\ Extended version}
%
%\titlerunning{Abbreviated paper title}
% If the paper title is too long for the running head, you can set
% an abbreviated paper title here
%
\author{Ying Li\inst{1,2}\orcidID{0000-0002-1367-9679} \and Patrick Lambrix\inst{1,2}\orcidID{0000-0002-9084-0470} }
%\author{}
%\institute{}
\institute{Link{\"o}ping University, Link{\"o}ping, Sweden  \and The Swedish e-Science Research Centre, Link{\"o}ping, Sweden}

\maketitle              % typeset the header of the contribution

\setcounter{footnote}{0}

\begin{abstract}
The quality of ontologies and their alignments is crucial for developing high-quality semantics-based applications. Traditional debugging techniques repair ontology networks by removing unwanted axioms and mappings, but may thereby remove consequences that are correct in the domain of the ontology network. 
In this paper we propose a framework for repairing ontology networks that deals with this issue. It defines basic operations such as debugging, weakening and completing. Further, it defines combination operators that reflect choices in how and when to use the basic operators, as well as choices regarding the autonomy level of the ontologies and alignments in the ontology network.
We show the influence of the combination operators on the quality of the repaired network and present an implemented tool. 
By using our framework together with existing algorithms for debugging, weakening and completing, we essentially provide a blueprint for extending previous work and systems. \footnote{This is a slightly revised and extended version of \cite{LL24}.}

%\keywords{First keyword  \and Second keyword \and Another keyword.}
\end{abstract}

\section{Introduction}

Ontologies and ontology networks, i.e., a set of ontologies connected through alignments,  are core components in application scenarios that involve searching, integrating, managing and extracting value from diverse sources of data at large scale. In the latter case they are the input for machine learning and data mining applications in diverse fields such as business analytics, health, crime analysis, and materials design. They are also the structural part of knowledge graphs which are used by, e.g., major data and database providers such as Google, Amazon, Meta, and Neo4j. 
The quality of ontologies, ontology networks and knowledge graphs is crucial for developing high-quality semantics-based applications. However, ensuring their quality, in particular regarding completeness (or coverage, all relevant information is modeled) and correctness (no wrong information is modeled), is a major challenge \cite{Pau16,NGJNPT19,Hog22,Che23,Lam23}.

Repairing ontology networks is a natural requirement for any application that would need to use several ontologies. We have, for instance, experience in a number of settings where an ontology network is repaired. A first example is the case of ontologies in the Ontology Alignment Evaluation Initiative (OAEI) where the ontologies to be aligned and a partial alignment were used to detect and repair defects \cite{LI13} (and the results were sent to the ontology owners). Another example is the development of modular ontologies where the modules and connections between the modules have been repaired (e.g., \cite{LALHPL24}). Further, we have worked with companies that used publicly available ontologies as well as concern-wide ontologies and where the network of these were repaired  (e.g., \cite{IBHL12}). 

The workflow for dealing with unwanted axioms in ontologies or networks consists of several steps including the detection and localization of the defects and the repairing. In this paper we assume we have detected defects and we now need to repair the ontologies and networks.
In the classical approaches the end result is a set of axioms to remove from the ontology or network that is obtained after detection and localization, and the repairing consists solely of removing
the suggested axioms. However, more knowledge than needed may be removed and approaches that alleviate the effect of removing unwanted axioms using weakening and completing have been proposed. In \cite{LL23} a framework for repairing $\el$ ontologies, based on the basic operations removing, weakening and completing, was proposed. Further, different combination operators were introduced that relate to choices that can be made when combining the basic operations (e.g., in which order to perform the operations, and when to update the ontologies).  
It was shown that the choice of combination has an influence on the amount of validation work by a domain expert and the completeness of the final ontology. It was also shown that earlier work on weakening (without completing) only considered one of the possible combinations. Similarly, earlier work on completing (without weakening) also considered only one of the possible combinations. The framework was extended with the basic operation debugging in \cite{LL24submitted}.

Our contributions in this paper relate to different ways in which we extend the framework in \cite{LL23,LL24submitted}  to deal with \textit{ontology networks}.
(i) We formalize the problem,  and show that new choices appear in addition to the choices for ontologies, when ontologies are connected with alignments  (Sect. \ref{sec-problem-definition}) (ii) We introduce different levels of autonomy for the ontologies and the alignments in the ontology network, reflecting the policies of the ontology and alignment owners regarding updating and computing for their ontologies and alignments (Sect. \ref{sec-problem-definition}). We define combination operators based on these autonomy levels and show the influences on the quality of the repair and the validation work (Sect. \ref{sec-combinations}). In general, there is a trade-off between the quality of the repaired ontologies on the one hand, and the amount of validation work for the oracle (e.g., domain experts) and the autonomy level for the ontologies and alignments on the other hand. We also give recommendations about which combinations to use in different cases, and we show that current systems for alignment repair (a special case of ontology network repair) use one particular combination of choices.
In Sect. \ref{sec-experiments} we give examples of repairing existing ontology networks and (iii) in Sect. \ref{sec-system} we briefly describe an implemented tool that allows users to make different choices for the combination operators.

%\textcolor{red}{statement about suppl material}

The only other work that discusses the combination of basic operations for repairing is \cite{LL23,LL24submitted} where the focus is on ontologies and the basic operations are removing, weakening and completing in \cite{LL23}, and additionally debugging in \cite{LL24submitted}. For the basic operations different approaches can be used. As this paper is not about possible approaches for the basic operations, but rather about their combinations, we mention these approaches in Sect. \ref{sec-basic-operations} and refer for an overview that compares these approaches to 
%the recent 
\cite{Lam23}. 
This paper essentially treats the basic operations of debugging, weakening and completing as black boxes and does not put requirements on the actual implementation of these operations. Therefore, by using our framework and the combination operators together with existing algorithms for debugging, weakening and completing, we essentially provide a blueprint for extending 
%our and other groups' 
previous work and systems.

\begin{table}[hbt!]
\begin{center}
\vspace{-0.2cm}
\caption{$\elb$ syntax and semantics. }
\begin{tabular}{|c|c|c|}

\hline
Name & Syntax & Semantics \\
\hline
\hline
top & $\top$ & $\Delta^{\mathcal I}$ \\
\hline
bottom & $\perp$ & $\emptyset$ \\
\hline
conjunction & $P \sqcap Q$ & $P^{\mathcal I} \cap Q^{\mathcal I}$ \\
\hline
existential restriction & $\exists r. P$ & $\{ x \in \Delta^{\mathcal I}$ $\mid$ $\exists y \in \Delta^{\mathcal I} :  $ \\
 &  &  $(x,y) \in r^{\mathcal I} \wedge y \in P^{\mathcal I}\}$\\
\hline
\hline
GCI & $P \sqsubseteq Q$ & $P^{\mathcal I} \subseteq Q^{\mathcal I}$\\
\hline
\end{tabular}
\end{center}
\label{tab:elb}
\end{table}
\vspace{-0.15cm}
\section{Preliminaries}
\label{sec-preliminaries}

\subsection{Description logics}
 
Description logics (DL) \cite{BCMNP03} are knowledge representation languages where concept descriptions are constructed inductively from a set $N_C$ of atomic concepts and a set $N_R$ of atomic roles and (possibly) a set $N_I$ of individual names. 
Different DLs allow for different constructors for defining complex concepts and roles. 
An interpretation $\mathcal I$ consists of a non-empty set $\Delta^{\mathcal I}$ and an interpretation function $\cdot^{\mathcal I}$ which assigns to each atomic concept $P \in N_C$ a subset $P^{\mathcal I} \subseteq \Delta^{\mathcal I}$, to each atomic role $r \in N_R$ a relation $r^{\mathcal I} \subseteq \Delta^{\mathcal I} \times \Delta^{\mathcal I}$, and to each individual name\footnote{As we do not deal with individuals in this paper, we do not use individuals in the later sections.} $i \in N_I$ an element $i^{\mathcal I} \in \Delta^{\mathcal I}$.
The interpretation function is straightforwardly extended to complex concepts. A TBox is a finite set of axioms which in $\elb$  are \emph{general concept inclusions} (GCIs). 
The syntax and semantics for $\elb$  are shown in Table \ref{tab:elb}.

An interpretation ${\mathcal I}$ is a \emph{model} of a TBox ${\mathcal T}$ if  for each GCI  in ${\mathcal T}$, the semantic conditions are satisfied.
We say that a TBox ${\mathcal T}$ is \emph{inconsistent} if there is no model for ${\mathcal T}$. Further, a
concept $P$ in a TBox ${\mathcal T}$ is \emph{unsatisfiable} if for all models ${\mathcal I}$ of ${\mathcal T}$ : $P^{\mathcal I}$ = $\emptyset$. We say that a TBox is \emph{incoherent} if it contains an unsatisfiable concept.
One of the main reasoning tasks for DLs is subsumption checking\footnote{Note that unsatisfiability checking in $\elb$ can be be performed using subsumption checking. A concept $P$ is unsatisfiable if  $P$ $\sqsubseteq$ $\perp$. Further, we can express that two concepts $P$ and $Q$ are disjoint by requiring that $P \sqcap Q \sqsubseteq \perp$.}
in which the problem is to decide for a TBox ${\mathcal T}$ and concepts $P$ and $Q$ whether ${\mathcal T}$  $\models P \sqsubseteq Q$, i.e., whether $P^{\mathcal I} \subseteq Q^{\mathcal I}$ for every model $\mathcal I$ of TBox ${\mathcal T}$. 
In this paper we update the TBox during the repairing and we always use subsumption with respect to the current TBox.

\subsection{Ontology networks}

In this paper we assume that ontologies are represented using DL TBoxes. 
An alignment between two ontologies is a set of mappings between the ontologies. A mapping between two ontologies is represented by $P$ $\sqsubseteq$ $Q$ where $P$ is a concept in the first ontology and $Q$ is a concept in the second ontology. Note that equivalence mappings (e.g., $P$ is equivalent to $Q$) are represented by two subsumption mappings ($P$ $\sqsubseteq$ $Q$ and $Q$ $\sqsubseteq$ $P$). Although we base our work and examples on $\elb$, the discussions hold for ontologies represented by DLs in general. An ontology network is a collection of ontologies together with their alignments and can be represented by a TBox (Def. \ref{def-ontology-network}).\footnote{We do not require that the ontologies in the network have the same signature.} In the remainder we use the term axiom for the axioms in the ontologies and the mappings. When we mean axioms in the ontologies, we will explicitly state this.

\begin{mydef}
Let ${\mathcal T}_1$, ...,${\mathcal T_n}$ be TBoxes representing ontologies ${\mathcal O_1}$, ..., ${\mathcal O_n}$, respectively. For $i, j \in [1..n]$ with $i < j$, let ${\mathcal A_{ij}}$ be an alignment between ontology ${\mathcal O_i}$ and ${\mathcal O_j}$. The network of the ontologies and their alignments is then represented by TBox ${\mathcal T}$ =  ($\bigcup_{i=1..n}$ ${\mathcal T}_i$)  $\cup$ ($\bigcup_{i,j=1..n, i<j}$ ${\mathcal A_{ij}}$). 
\label{def-ontology-network}
\end{mydef}

Our aim is to find repairs that remove as much wrong knowledge and add as much correct knowledge (back) to our ontology network as possible. Therefore, we use the preference relations \textit{more complete} and \textit{less incorrect} between TBoxes (Defs. \ref{def-preferences-complete} and \ref{def-preferences-incorrect}) that formalize these intuitions \cite{Lam23}. Further, if a TBox representing an ontology (network) is more complete/less incorrect than the Tbox of an other ontology (network), then we say that the first ontology (network) is more complete/less incorrect than the second ontology (network). The definitions assume the existence of an oracle (representing a domain expert) that, when given an axiom, can answer whether this axiom is correct or wrong in the domain of interest of the ontology network.

\begin{mydef} (more complete)
\label{def-preferences-complete}
TBox ${\mathcal T}_1$ is \emph{more complete} than TBox ${\mathcal T}_2$ (or ${\mathcal T}_2$ is \emph{less complete} than  ${\mathcal T}_1$) according to oracle $Or$ iff 
$(\forall \psi: ({\mathcal T}_2  \models \psi \wedge ~Or(\psi) = true)    \rightarrow  {\mathcal T}_1 \models \psi)) 
\wedge (\exists \psi:  Or(\psi) = true ~\wedge {\mathcal T}_1  \models \psi \wedge {\mathcal T}_2 \not\models \psi)$. They are \emph{equally complete} iff 
$\forall \psi: Or(\psi) = true  \rightarrow 
({\mathcal T}_1 \models \psi  \leftrightarrow  {\mathcal T}_2 \models \psi)$
\end{mydef}
\begin{mydef} (less incorrect)
\label{def-preferences-incorrect}
TBox ${\mathcal T}_1$ is \emph{less incorrect} than TBox ${\mathcal T}_2$ (${\mathcal T}_2$ is \emph{more incorrect} than  ${\mathcal T}_1$) according to oracle $Or$ iff 
$(\forall \psi: ({\mathcal T}_1 \models \psi \wedge Or(\psi) = false )   \rightarrow  {\mathcal T}_2 \models \psi)) 
\wedge (\exists \psi:  Or(\psi) = false ~\wedge {\mathcal T}_1 \not\models \psi  \wedge {\mathcal T}_2 \models \psi)$. ${\mathcal T}_1$  and ${\mathcal T}_2$ are \emph{equally incorrect} iff 
$\forall \psi: Or(\psi) = false  \rightarrow 
({\mathcal T}_1 \models \psi  \leftrightarrow  {\mathcal T}_2 \models \psi)$. 
\end{mydef}

\subsection{Basic operations}
\label{sec-basic-operations}

To repair ontologies, algorithms can be developed using a number of basic operations such as debugging, removing, weakening and completing, and combining these in different ways \cite{LL23,LL24submitted}. In this paper we use variants of these operations to repair ontology networks. 

\begin{figure}[t]
\begin{center}
\vspace{-0.25cm}
{\includegraphics[width=0.14\textwidth]{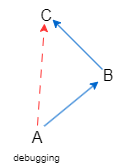}}  
{\includegraphics[scale=0.27]{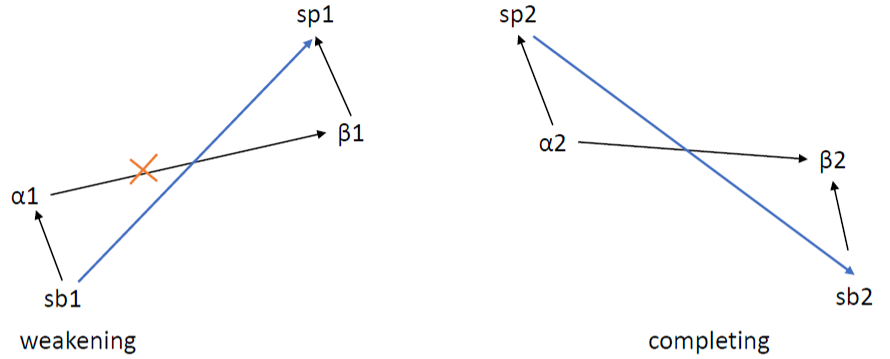}}
\caption{Debugging, weakening and completing.} 
\label{fig-weakening-completing}    
\vspace{-0.65cm} 
\end{center}
\end{figure}

Given a set of wrong axioms $W$, \textit{debugging} aims to find a set of wrong \texttt{asserted} axioms $D$ that when all axioms in $D$ are removed from the network, the axioms in $W$ cannot be derived anymore.
%(by contrast with the removing in \cite{LL23}, during debugging, the justifications of each wrong axiom will be computed and then the domain expert will validate the asserted wrong axioms to remove from the ontology). 
Many debugging approaches have been proposed  (e.g., \cite{SC03,Sch05,KPSC06,MLBP06,SHCvH07,KPHS07,LB10,JGHZ11,MMV11,NRG12,SFFR12,FMVN13,AMIMPM16,KS18,RS18,SRS18,MAKLBD23}, overview in \cite{Lam23})\footnote{Many of these approaches use axiom pinpointing \cite{Pen20}. Further, there are other approaches that take ABoxes into account.}. A basic approach is based on the computation of justifications for the wrong axioms and then computing a Hitting set over the set of justifications. As an example, in Fig. \ref{fig-weakening-completing} derived wrong axiom A $\sqsubseteq$ C needs to be removed. This can be done by removing asserted axiom A $\sqsubseteq$ B or asserted axiom B $\sqsubseteq$ C.
\textit{Removing} deletes all the wrong asserted axioms in a given set $D$ from the ontology network. Removing makes a network less or equally incorrect than it was before the operation.
Given an axiom, \textit{weakening} aims to find other axioms that are weaker than the given axiom, i.e., the given axiom logically implies the other axioms within the network. For the repairing this means that a wrong axiom  $\alpha$ $\sqsubseteq$ $\beta$ can be replaced by a correct weaker axiom  $sb$ $\sqsubseteq$ $sp$ such that $sb$ is a sub-concept  of $\alpha$ and $sp$ is a super-concept  of $\beta$, thereby mitigating the effect of removing the wrong axiom (Fig. \ref{fig-weakening-completing}). Algorithms for weakening have been provided in e.g., \cite{TCGPPK18,BKNP18,CGKPRT20,LL23}.
\textit{Completing} aims to find correct axioms that are not derivable from the ontology yet and that would make a given axiom derivable. 
For a given axiom $\alpha$ $\sqsubseteq$ $\beta$, it finds correct axioms $sp$ $\sqsubseteq$ $sb$ such that $sp$ is a super-concept of $\alpha$ and $sb$ is a sub-concept of $\beta$ (Fig. \ref{fig-weakening-completing}). This means that if $sp$ $\sqsubseteq$ $sb$ is added to ${\mathcal T}$, then $\alpha$ $\sqsubseteq$ $\beta$ would be derivable. Completing is performed on correct axioms, and in repairing, it is applied to weakened axioms. Completing algorithms are proposed in, e.g., \cite{LDI12,WDL14,DWM17,HKTW22,LL23}.
Note that weakening and completing are dual operations where the former finds weaker axioms and the latter stronger axioms. Both these operations make an ontology network more or equally complete.
%than it was before the operation.

\begin{figure}[t] 
\begin{center}
 \vspace{-0.25cm}
\subfigure[Debugging]{\includegraphics[width=0.3\textwidth]{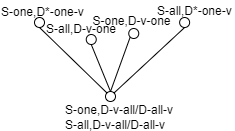}}\label{a}
\subfigure[Removing]{\includegraphics[width=0.2\textwidth]{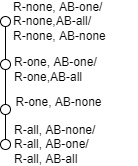}}\label{b}
\subfigure[Weakening]{\includegraphics[width=0.2\textwidth]{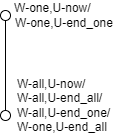}} \label{c}
\subfigure[Completing]{\includegraphics[width=0.2\textwidth]{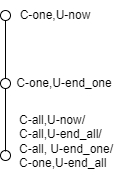}}\label{d} 
\caption{
Hasse diagrams (a from \cite{LL24,LL24submitted} and b-d from \cite{LL23}).
(a) selecting and debugging;
(b) removing and adding back wrong axioms; 
(c) weakening and updating;  
(d) completing and updating.
}
\label{fig-lattice}   
\end{center}
\vspace{-0.4cm}
\end{figure}

These basic operations can be combined in different ways and there are choices to be made in terms of, e.g., the order in which the operations are performed, the order in which the axioms are processed, whether one axiom is dealt with at a time or all at once, and when the TBox is updated.  For removing, weakening and completing, the different combinations were classified in Hasse diagrams in \cite{LL23}, and for debugging in \cite{LL24submitted} (Fig. \ref{fig-lattice}). Explanations for removing, weakening and completing, and for debugging are from \cite{LL23} and \cite{LL24submitted}, respectively and are given in the appendix. The proofs of the Hasse diagrams are given in the appendix.
Note that the discussions and proofs for all Hasse diagrams in this paper hold for ontologies and ontology networks represented by DL TBoxes in general, and thus the results regarding the combination operators are language agnostic.\footnote{When the framework is instantiated to create actual systems, then choices need to be made regarding which algorithms to use for the different basic operations, and the choice of these algorithms will, of course, also influence the completeness and correctness of an ontology or ontology network (but this is not the topic of this paper).}

In general, operations higher up in the diagrams use more (but also more possibly wrong) information during the computations and lead to more (or equally) complete ontologies, more (or equally) incorrect ontologies as well as more validation work for the domain expert. %\footnote{Debugging and removing aim to reduce the incorrectness. However, when removing wrong knowledge, correct derived knowledge may also be removed and thus the network may become less complete. Similarly, weakening and completing aim to make the network more complete. However, by adding correct knowledge, this knowledge may, together with existing wrong knowledge, allow deriving more wrong knowledge as well, thus making the network more incorrect.} 
Algorithms for repairing can be defined by which of the operators are used and in which order. These algorithms can then be compared using the Hasse diagrams.
In general, if the sequence of operators for one algorithm can be transformed into the sequence of operators of a second algorithm, by replacing some operators of the first algorithm using operators higher up in the  Hasse diagrams in  Fig. \ref{fig-lattice}, then the ontologies repaired using the second algorithm are more or equally complete and incorrect than the ontologies repaired using the first algorithm.
The work in \cite{LL23} was the first to identify these different combinations. It also showed that previous work on combining removing with weakening used the combination (R-one, AB-none) with (W-one, U-now). Previous work on completing used (C-one, U-end-all). There was no work that used other combinations and no work that combined all basic operations.

For the examples in this paper we have used Algorithm \ref{algorithmexample} as a basis for discussing repairing of ontology networks. It uses a particular combination of debugging, removing, weakening and completing with specific algorithms for the basic operations. Our discussion regarding the choices for {\it ontology networks} (Sect. \ref{sec-problem-definition}) would still hold if we used other algorithms for the basic operations, or different combination operators for the basic operators.

 %\vspace{-0.1cm}
\begin{algorithm}[h]
\caption{Generate the justifications of each wrong axiom and validate all wrong asserted axioms from the generated justifications,  weaken all wrong asserted axioms, complete all weakened axioms, add all completed axioms and remove all wrong asserted axioms at the end (S-one,D-v-all/R-none,AB-none/W-all,U-end\_all/C-all,U-end\_all).}%Algorithm9\_V1 
     \hspace*{\algorithmicindent}\textbf{Input}: TBox ${\cal T}$,  Oracle Or, set of unwanted axioms $W$ 
     \\
   \hspace*{\algorithmicindent}\textbf{Output}: A repaired TBox
	\begin{algorithmic}[1]
        \State $D\leftarrow$ $\emptyset$
        \ForEach {$\alpha$ $\sqsubseteq$ $\beta$ $\in$ $W$}
         \ForEach {  $J\in$ Justifications$({\alpha \sqsubseteq \beta})$}
         %\State $D\leftarrow$ $D$ $\cup$ \{ Validate-one-wrong-axiom$(J,Or)$ \}
         \ForEach {  $axiom \in$ $J$}
                \If {$Validate\_axiom(axiom,Or)= false$ }
             \State $D \leftarrow$ $D $ $\cup$ \{$axiom$\}
                \EndIf
                \EndFor  
         \EndFor
          \EndFor
	\ForEach {$\alpha$ $\sqsubseteq$ $\beta$ $\in$ $D$}
	%\State ${\cal T}_r$ $\leftarrow$ Remove-axioms(${\cal T}$, $\{\alpha \sqsubseteq\beta\}$)
   \State $w_{\alpha \sqsubseteq \beta}\leftarrow$ weakened-axiom-set($\alpha \sqsubseteq \beta$, ${\cal T}, Or)$
   \EndFor
        \State $c_{\alpha \sqsubseteq \beta}\leftarrow$ $\emptyset$
	\ForEach {$\alpha$ $\sqsubseteq$ $\beta$ $\in$ $D$}
   	\ForEach {$sb$ $\sqsubseteq$ $sp$ $\in$ $w_{\alpha \sqsubseteq \beta}$}
   \State $c_{sb \sqsubseteq sp}\leftarrow$ completed-axiom-set($sb \sqsubseteq sp$,${\cal T},Or)$
   \State $c_{\alpha \sqsubseteq \beta}\leftarrow$ $c_{\alpha \sqsubseteq \beta}$ $\cup$ $c_{sb \sqsubseteq sp}$
   \EndFor
   
  \EndFor
  \State ${\cal T}_r$ $\leftarrow$  Add-axioms(${\cal T}$,$\bigcup_{\alpha \sqsubseteq\beta}$ $c_{\alpha \sqsubseteq \beta}$)
  \State ${\cal T}_r$ $\leftarrow$ Remove-axioms(${\cal T}_r$,$D$)
   \State \Return ${\cal T}_r$
	\end{algorithmic}
	\label{algorithmexample}
\end{algorithm}
\vspace{-0.1cm}

\section{Repairing ontology networks - problem definition}
\label{sec-problem-definition}

In this section we define the repairing problem for ontology networks. Def. \ref{def-repair} is an extension of the definition of repair for single ontologies as defined in \cite{LL23}.
We are given a set of wrong axioms $W$ that we want to remove from the ontology network, and when they are removed, they cannot be derived from the TBox representing the ontology network anymore. These axioms in $W$ can be axioms in the ontologies or mappings. Further, to guarantee a high level of quality of the ontology (i.e., so that no correct information is removed or no incorrect information is added), domain expert validation is a necessity (e.g., \cite{PFSC13,LL23}).
Therefore, we assume an oracle (representing a domain expert) that, when given an axiom, can answer whether this axiom is correct or wrong in the domain of interest of the ontology network. A repair ($A$, $D$) for the ontology network given the TBox ${\mathcal T}$, oracle $Or$, and a set of wrong axioms $W$, is a tuple containing two sets: a set $A$ of axioms that are correct according to the oracle and should be added to the TBox, and a set $D$ of asserted axioms that are not correct according to the oracle and should be removed from the TBox. We require that when the axioms in $D$ are removed and the axioms in $A$ are added, the wrong axioms in $W$ cannot be derived anymore.

\begin{mydef} (Repair)
Let TBox ${\mathcal T}$ =  ($\bigcup_{i=1..n}$ ${\mathcal T}_i$)  $\cup$ ($\bigcup_{i,j=1..n, i<j}$ ${\mathcal A_{ij}}$) represent a network of ontologies ${\mathcal O_{i}}$ represented by TBoxes ${\mathcal T_{i}}$, and their alignments ${\mathcal A_{ij}}$.
Let $Or$ be an oracle that given a TBox axiom returns true or false.
Let $W$ be a finite set of TBox axioms in ${\mathcal T}$ such that $\forall$ $\psi$ $\in$ $W$: $Or$($\psi$) = false.
Then, a repair for Debug-Problem DP$({\mathcal T}, Or, W)$ is 
a tuple ($A$, $D$) where $A$ and $D $ are finite sets of TBox axioms such that \\
(i) $\forall$ $\psi$ $\in$ $A$: $Or$($\psi$) = true;\\
(ii) $D$ is a finite set of \textit{asserted} axioms in ${\mathcal T}$; \\
(iii) $\forall$ $\psi$ $\in$ $D$: $Or$($\psi$) = false;\\
(iv) $\forall$ $\psi$ $\in$ $W$: $({\mathcal T} \cup A) \setminus D$ $\not \models$ $\psi$.
\label{def-repair}
\end{mydef}

From a theoretical point of view, as we have represented the ontology network as a TBox, and previous work has represented ontologies as TBoxes, we could reuse the algorithms and approaches in, e.g., \cite{LL23,LL24submitted}. However, from a practical point of view, the situation is more complex. When working with single ontologies, it is in the interest and mandate of the ontology owners to repair their ontologies. However, when ontologies are connected in a network, the ontology and alignment owners may want to retain different levels of autonomy and not necessarily allow others to change or propose changes to their ontologies and alignments. Also, computation time and validation work for repairing may be issues. In this case, computation time and validation work may be lower within an ontology or alignment than for the whole network. In this paper we discuss three levels of autonomy and show the influence of these different choices.

The first level consists of the cases 'O' (ontology) and 'M' (mappings). 'O' represents the choice where ontologies are completely autonomous. Essentially, this means that ontologies act on their own regarding detection and repairing of defects. $W$ in Def. \ref{def-repair} contains only axioms in the ontology, only the axioms within the ontology itself can be used for the computation of repairs, and solutions can only include axioms in the ontology itself. 
A dual case is 'M' where the owners of an alignment are autonomous. In this case $W$ contains only mappings, and the alignment is repaired using only the mappings in the alignment.
The second level consists of the cases 'MO' (materialized ontology) and 'MM' (materialized mappings). 'MO' uses the network to derive new axioms within the ontology and materializes these axioms.\footnote{We note that a choice may be made regarding which axioms to materialize. For taxonomies it may be feasible to materialize all. However, for $\elb$ there is an infinite number of derivable axioms. In our experiments we restrict the space to axioms with concepts at the left- and right-hand side with non-nested operators (SCC in \cite{LL23}).} $W$ contains only axioms in the ontology. For the computation of repairs the materialized ontology is used, and solutions contain only axioms within the ontology. This level accepts the fact that the network provides more knowledge about the own ontology than is represented by the ontology itself, and accepts this knowledge, but it does not use the network in repairing. 
'MM', the dual case for mappings, computes derived mappings for the alignment, but then acts autonomously. $W$ contains only mappings.
The third level 'ON' (ontology network) considers the ontologies and alignments as integral parts of the network, uses the full network for the computation of repairs, and repairs defects using axioms within all ontologies and alignments. $W$ can contain ontology axioms and mappings.
We note that different ontology and alignment owners may make different choices regarding their level of autonomy.

\begin{figure}[ht!] 
\begin{center}
 \vspace{-0.25cm}
\subfigure[]{\includegraphics[width=0.3 \textwidth]{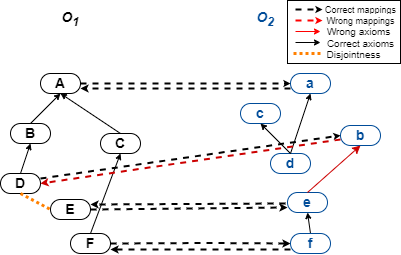}}\label{a}
\subfigure[]{\includegraphics[width=0.52\textwidth]{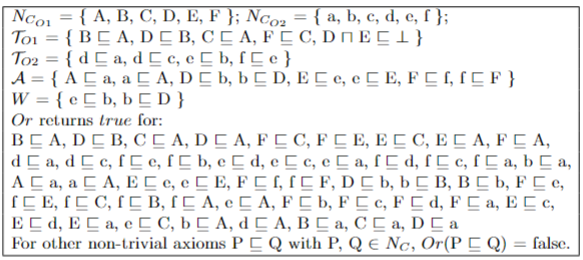}}\label{b}
\caption{The network contains unsatisfiable concepts E, F, e, and f. 
Asserted axiom e $\sqsubseteq$ b in $O_{2}$, and mapping b $\sqsubseteq$ D between $O_{2}$ and $O_{1}$ lead to network incoherence.  }
\label{example-network}  
\end{center} \vspace{-0.6cm}
\end{figure}

These choices have an influence on the detection and repairing of effects. We give an example for the detection here, and focus on the repairing in further sections.
For the network in Fig. \ref{example-network} we have the following situation. 
Using the full network (level 'ON'), we can derive that concepts E, F, e, and f are unsatisfiable.\footnote{E unsatisfiable because D $\sqcap$ E $\sqsubseteq$ $\perp$, E $\sqsubseteq$ e, e $\sqsubseteq$ b, b $\sqsubseteq$ D. e unsatisfiable because E unsatisfiable and e $\sqsubseteq$ E. f unsatisfiable because e unsatisfiable and f $\sqsubseteq$ e. F unsatisfiable because f unsatisfiable and F $\sqsubseteq$ f.} 
Using level 'MO' for both ontologies, we materialize E $\sqsubseteq$ D and F $\sqsubseteq$ E in the first ontology, and e $\sqcap$ b $\sqsubseteq$ $\perp$ in the second ontology. This allows us to obtain the same unsatisfiable concepts as in the 'ON' level (although the repairing will be different).
When using level 'O' for both ontologies, we cannot detect any unsatisfiable concepts.

In general, there is a trade-off between the quality of the repaired ontologies on the one hand, and the amount of validation work for the oracle and the autonomy level for the ontologies and alignments on the other hand. In the next sections we define and exemplify these notions.

\section{Repairing ontology networks - Combination operators}
\label{sec-combinations}

During the repairing process different levels of autonomy can be used at different stages. The choice has an influence on the quality of the repair. Here we show this influence for different stages.

\begin{figure}[ht!] 
\begin{center}
 \vspace{-0.2cm}
\subfigure[]{\includegraphics[width=0.15 \textwidth]{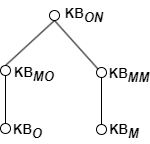}}\label{a}
\subfigure[]{\includegraphics[width=0.16\textwidth]{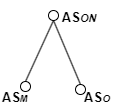}}\label{b}
\caption{Hasse diagrams regarding (a) the use of different knowledge bases during repairing (KB); (b) different restrictions on the add set (AS);  }
\label{Hasse-network}   
\end{center} \vspace{-0.2cm}
\end{figure}

\begin{table}[h]
\begin{center}
 \vspace{-0.3cm}
\caption{\label{tab:choices-overview}Choices for background knowledge (KB) and add sets (AS).}
\begin{tabular}{|p{1.2cm}|l|}
\hline
Choices & Description \\
\hline
KB$_{ON}$&Use the whole ontology network as the background knowledge base to compute \\&the sub/sup-concepts when weakening and completing an axiom. \\
KB$_{MO}$&Materialize each ontology in the network. Disconnect each ontology from the \\& ontology network. Use only the axioms within the materialized ontologies to \\&compute the sub/sup-concepts when weakening and completing an axiom. \\
KB$_{O}$&Disconnect each ontology from the ontology network. Use the axioms within the\\& respective ontologies to compute the sub/sup-concepts when weakening and \\& completing an axiom. \\
KB$_{MM}$&Materialize each alignment in the network. Disconnect each alignment from the \\& ontology network. Use only the mappings within the materialized alignments to \\&compute the sub/sup-concepts when weakening and completing an axiom. \\
KB$_{M}$&Disconnect each alignment from the ontology network. Use the mappings within \\&the respective alignments to compute the sub/sup-concepts when weakening and \\& completing an axiom. \\
\hline
\hline
AS$_{ON}$&Add all axioms.\\
AS$_{O}$ &Add only axioms within the respective ontologies.\\
AS$_{M}$ &Add only mappings between the ontologies. \\
\hline
\end{tabular}
\end{center}
\end{table}
\vspace{-0.4cm}

\subsection{Use of background}

As discussed earlier, we can use different autonomy levels during the computation of repairs. In our algorithms, this is reflected by which TBox is used. We can combine particular choices for debugging, removing, weakening and completing algorithms with different autonomy levels. As an example, assume we have made the choices for the basic operations as in Algorithm 1. Then, for autonomy level 'ON' we can use Algorithm 1 with as input the TBox representing the whole network. For levels 'O' and 'MO' we can use Algorithm 1 repeatedly using each of the TBoxes of the ontologies or materialized ontologies, respectively. (Another possibility is to use as input the union of these TBoxes, i.e., there will be no axioms representing mappings in the network.) For 'M' and 'MM' we use the alignments or materialized alignments as TBoxes. The choices are explained in Table \ref{tab:choices-overview} where KB stands for 'knowledge base'. Note that the choices can be performed on each of the basic operations. For instance, it is possible to debug using 'ON' and then weaken and complete using 'MO'. We can organize the choices in a Hasse diagram as in Fig. \ref{Hasse-network}(a). The proofs for the Hasse diagram are given in the appendix.  In general, using more background knowledge leads to more (or equally) complete and more (or equally) incorrect networks, and more validation work by the oracle.

As an example, assume we have used 'ON in the debugging step where we validate all axioms in the justifications, for the network in Fig. \ref{example-network}. This results in wrong axioms e $\sqsubseteq$ b and b $\sqsubseteq$ D. We now look at the weakening step for the different autonomy levels. When weakening e $\sqsubseteq$ b in the second ontology, we use sub-concepts of e and super-concepts of b to find weakened axioms. These sets are different for different autonomy levels. We have Sub(e,KB$_{O}$)=\{e, f\}, Sub(e,KB$_{MO}$)=\{e, f\}, Sub(e,KB$_{ON}$)=\{e, f, E, F\}, and Sup(b,KB$_{O}$)=\{b\}, Sup(b,KB$_{MO}$)=\{b, a\}, Sup(b,KB$_{ON}$)=\{b, a, D, B, A\}. Therefore, there are 2, 4 and 20 candidates for the weakened axioms for 'O', 'MO' and 'ON', respectively. After validation we have as weakened axioms f $\sqsubseteq$ b for 'O', f $\sqsubseteq$ b  and  e $\sqsubseteq$ a for 'MO', and  f $\sqsubseteq$ b,  
e $\sqsubseteq$ a, and E $\sqsubseteq$ A for 'ON' (Table \ref{example-result1}). As expected from the Hasse diagram, 'ON' leads to the most complete network, followed by 'MO' and then 'O'. 
When weakening the wrong mapping b $\sqsubseteq$ D, we can choose between 'M', 'MM' and 'ON'. 
We have Sub(b,KB$_{M}$)=\{b, D\}, Sub(b,KB$_{MM}$)=Sub(b,KB$_{ON}$)=\{e, f, b, D, E, F\}, Sup(D,KB$_{M}$)=\{D, b\}, and Sup(D,KB$_{MM}$)=Sup(D,KB$_{ON}$)=\{D, A, B, b, a\}.\footnote{All algorithms for debugging, weakening and completing use heuristics when computing the sub-concepts and super-concepts to prune the search space of solutions \cite{Lam23,LL23}. Also for the networks different heuristics can be used. For instance, when computing the sub- and super-concepts for the concepts in mappings, one heuristic is to only allow sub- and super-concepts in the same ontology. In the example we have allowed sub- and super-concepts to belong to other ontologies as long as they are involved in mappings.}
%The sub-concepts of b are \{b, D\} for 'M' and \{e, f, b, D, E, F\} for 'MM' and 'ON'.\footnote{All algorithms for debugging, weakening and completing use heuristics when computing the sub-concepts and super-concepts to prune the search space of solutions \cite{Lam23,LL23}. Also for the networks different heuristics can be used. For instance, when computing the sub- and super-concepts for the concepts in mappings, one heuristic is to only allow sub- and super-concepts in the same ontology. In the example we have allowed sub- and super-concepts to belong to other ontologies as long as they are involved in mappings.}
%The super-concepts for D are \{D, b\} for 'M' and \{D, A, B, b, a\} for 'MM' and 'ON'. 
For 'M' there are no weakened axioms. For 'MM' we find b $\sqsubseteq$ B. For 'ON' we find b $\sqsubseteq$ B and e $\sqsubseteq$ a. The latter includes both mappings and ontology axioms.
Table \ref{example-result1} also shows the results after completing the weakened axioms  using the same autonomy level as for weakening. The network is repaired by removing the wrong axioms and adding the completed axioms.

\begin{table}[H]
\begin{center}
\vspace{-0.2cm} 
%\begin{tabular}{|l|l|l|}
\caption{Weakening and completing the wrong axioms e $\sqsubseteq$ b and b $\sqsubseteq$ D.}
\begin{tabular}{|p{2.5cm}|p{0.98cm}|p{0.98cm}|p{0.98cm}|p{0.03cm}|p{0.98cm}|p{0.98cm}|p{0.98cm}|}
\hline
Background  & KB$_O$ &KB$_{MO}$ & KB$_{ON}$ && KB$_M$ & KB$_{MM}$& KB$_{ON}$\\  
knowledge base & e $\sqsubseteq$ b &e $\sqsubseteq$ b & e $\sqsubseteq$ b &&b $\sqsubseteq$ D&b $\sqsubseteq$ D & b $\sqsubseteq$ D\\  
\hline
$\mid$Sub($\alpha$,${\mathcal T}$)$\mid$&2&2&4 & &2&6&6\\
$\mid$Sup($\beta$,${\mathcal T}$)$\mid$&1&2&5 & &2&5&5\\\hline
Weakened &f $\sqsubseteq$ b& f $\sqsubseteq$ b,   e $\sqsubseteq$ a & f $\sqsubseteq$ b,  
e $\sqsubseteq$ a, 
E $\sqsubseteq$ A & 
& & b $\sqsubseteq$ B& b $\sqsubseteq$ B, e $\sqsubseteq$ a \\\hline
$\mid$Sup($\alpha$,${\mathcal T}$)$\mid$&2&4 3 &10 7 7 & & &5&5 7 \\
$\mid$Sub($\beta$,${\mathcal T}$)$\mid$&1&3 5 &6 11 11 & & &7&7 11 \\\hline
Completed & f $\sqsubseteq$ b& f $\sqsubseteq$ b,  
e $\sqsubseteq$ d&f $\sqsubseteq$ b, %F $\sqsubseteq$ E,  
e $\sqsubseteq$ d,
E $\sqsubseteq$ C%, b $\sqsubseteq$ a 
& &  & b $\sqsubseteq$ B, B $\sqsubseteq$ b &b $\sqsubseteq$ B, B $\sqsubseteq$ b 
\\\hline

\end{tabular}
\vspace{-0.55cm} 
\label{example-result1} 
\end{center}
\end{table}

\subsection{Add sets}

Another stage where the choice of autonomy level influences the quality of the repair is when deciding what repairing solutions to retain for the final repair, i.e. $A$ in Def. \ref{def-repair}. The choices are summarized in Table \ref{tab:choices-overview} under 'AS' (Add Set).
Choice AS$_{ON}$ is the most general case and allows all kinds of axioms to be added to the network. This means that regardless which choice was used during the computation of the repairs, these repairs can be used without any adaptions. Choice AS$_{O}$ only allows to add axioms within ontologies. This represents the case where ontology owners only focus on repairing their own ontologies. Similarly, choice AS$_{M}$ only allows to add mappings and represents the case where alignment owners only focus on repairing their own mappings. In AS$_{O}$ and AS$_{M}$, not all repairing suggestions may be retained for the final solution. Therefore, when 'ON' was used during the computation of repairs and 'O' or 'M' is used during this stage, this may lead to a lower level of completeness for the network than if 'ON' is used for this stage. 
The Hasse diagram for these choices is given in Fig. \ref{Hasse-network}(b) and the proofs 
%for the Hasse diagram
are given in the appendix.

\subsection{Finalizing}

In the case of 'ON' the repairing solutions may contain axioms in different ontologies as well as mappings. In this case the ontology and alignment owners may decide to materialize the knowledge derived from the network which regards their ontology or alignment. From the network point of view the network does not change logically as the same axioms can be derived. From the ontology or alignment point of view, the difference appears when they are disconnected from the network. The materialized versions are more (or equally) complete. 

\subsection{Discussion}

When the network is owned by one entity (e.g., in the case of modular ontologies with mappings) or there is a tight cooperation between the owners of the individual parts of the network (e.g., in a consortium such as EMMO (\url{https://github.com/emmo-repo/EMMO}) or OBO (\url{https://obofoundry.org/})), and computation time and domain expert staff is not an issue, the recommended combination is to use (KB$_{ON}$, AS$_{ON}$, final materialization). This ensures the highest level of completeness for the network as well as for the ontologies and alignments when they are disconnected from the network.

The least complete networks come from the combinations (KB$_{O}$, AS$_{O}$) and (KB$_{M}$, AS$_{M}$). Essentially, this means that mappings between the ontologies exist, but the network is not taken into account at all during repairing. The repairs are the same as when repairing is done without the network, and this seems to be a common case in practice.

When ontology or alignment owners require full control over the computation resources, while still taking into account the knowledge in the network, then the combinations with KB$_{MO}$ and KB$_{MM}$ may be a good choice. Maintaining full control about what is added leads to choices with AS$_{O}$ and AS$_{M}$.

In the ontology alignment field there are alignment systems that also repair the alignment. The current systems require that the repair only contains mappings and thus use AS$_{M}$. Examples of such systems are ALCOMO \cite{Meilicke11}, LogMap \cite{JG11,JGZH12}, AgreementMakerLight \cite{SFPC15}. Regarding the background knowledge, they all use the choice KB$_{ON}$. RaDON \cite{JHQHS09} is a system focusing on network repair, but makes the same choices as the alignment systems with repair functionality.

%\textcolor{red}{Check articles where part of the ontology is fixed}
%Interactive ontology revision and ontodebug? 
%Adding an example about showing that keeping wrong knowledge in the ontology during repairing can derive new correct axiom?

\section{Experiments}
\label{sec-experiments}

%\textcolor{red}{Need some results here?}
% maybe put the result of the example network here. We have the table3 in section 4
As examples of the use of different choices for the combination operators, we performed experiments on 5 ontology networks. The ontologies (ekaw, sigkdd, iasted, cmt) used in these networks are from the conference track of the OAEI\footnote{http://oaei.ontologymatching.org/2023/conference/index.html}. We have used the parts of these ontologies that are expressible in $\elb$ in the sense that we removed the parts of axioms that used constructors not in $\elb$. 
%The mappings between two ontologies are rewritten to 2 sub-axioms, i.e., using axioms $\alpha \sqsubseteq \beta$ and $\beta \sqsubseteq \alpha$ to replace the mapping $\alpha = \beta$. 
We introduced wrong axioms in the ontologies and mappings between ontologies by replacing existing axioms with axioms where the left-hand or right-hand side concepts of the existing axioms were changed. Further, we also flagged some existing axioms as wrong in our full experiment set. All axioms were validated manually. The characteristics of the ontologies and the wrong axioms are shown in the supplemental material.

We repaired the networks using all choices regarding the use of background knowledge together with Algorithm 1. To repair ontology axioms we used KB$_{O}$, KB$_{MO}$ and KB$_{ON}$. To repair mappings we used KB$_{M}$, KB$_{MM}$ and KB$_{ON}$. We computed the sizes of the sets of sub-concepts and super-concepts used in weakeing and completing. The sizes of these sets reflect the number of axioms that need to be validated by the oracle. We note, however, that using the visualization in our system (see below) these sets are shown together and thus the validation of many axioms can be done at the same time.

The full results for the experiments are shown in the appendix.
%Tables \ref{ekaw-sigkdd}-\ref{iasted-sigkdd} in the appendix.
In the experiments, the number of axioms to be validated for KB$_{ON}$ is between 3 and 5 times higher than for KB$_{O}$/KB$_{M}$. The validation work for the materialized versions KB$_{MO}$/KB$_{MM}$ is between 6\% and 40\% higher than for KB$_{O}$/KB$_{M}$. For all networks there were axioms for which the repairing with an operator higher up in the Hasse diagram led to a strictly more complete ontology network.

\section{Implemented system}
\label{sec-system}

We implemented a Java-based system which extends the $\el$ version of the RepOSE system (\cite{WDL14,LWD15,LL23}) with full debugging, weakening and completing for $\elb$ ontology network repairing. The system allows the user to choose different combinations, thereby giving a choice in the trade-off between validation work, incorrectness and completeness. 

The system uses an interactive way to repair the ontology network. 
%input
%If the user doesn't know the wrong axioms, the user can ask for the incoherence checking. After that if the ontology network is incoherent, the system will compute the justifications of these unsatisfiable concepts. If the user knows the wrong axioms, then, the user can input these wrong axioms and ask the system to compute the justifications of these wrong axioms. 
It takes as input the ontologies and alignments in the network as well as a set of wrong axioms. It is also possible to not give a set of wrong axioms but let the system deduce unsatisfiable concepts in the network. As basic algorithms, we have used the black-box algorithm in \cite{KPHS07} for debugging, and the weakening and completing algorithms in \cite{LL23}. With respect to the combination operators in the Hasse diagrams, the system supports (S-one, D-v-all) and (S-one, D-one-v) thereby providing the choice to validate all axioms in the justifications or to validate Hitting sets. Regarding weakening the combinations are (R-none,AB-none/W-all,U-end\_all), (R-all,AB-none/W-all,U-end\_all) and (R-one,AB-one/W-one,U-end\_all).  The first combination does not remove wrong axioms during the computation, weakens all axioms at once and updates at the end. The second removes all wrong axioms before the computation, weakens all axioms at once and updates at the end. The third removes wrong axioms one at a time during the computation and puts them back before dealing with the next wrong axiom. It weakens axioms one at a time and updates at the end.
For completing, the choices are (C-all,U-end\_all) and (C-one,U-end\_one). The first combination completes all weakened axioms at once and updates at the end. The second completes the weakened axioms one at a time and updates the ontology after the weakened axiom set is handled for each wrong axiom.
Regarding the choices for the combinations for the network, all choices for KB are implemented.
%KB$_{ON}$, KB$_{MO}$, KB$_{MO}$, KB$_{O}$ and KB$_{M}$, they are all implemented.
%\textcolor{red}{would it be easy to implement MM?}
The AS choices are supported using visual clues. The concepts in the axioms to be validated are labeled with the ontology source, such that the user can distinguish mappings from axioms within the ontologies. We also use different colors to represent the concepts which belong to different ontologies.
%. i.e. the concepts which belong to the first ontology will be blue and the concepts which belong to the second ontology are black ones.
At the appropriate times the system shows the different combinations that can be chosen and the user can select the desired choice.

\begin{figure}[htb!] 
\begin{center}
 \vspace{-0.2cm}
\subfigure[]{\includegraphics[width=0.41\textwidth]{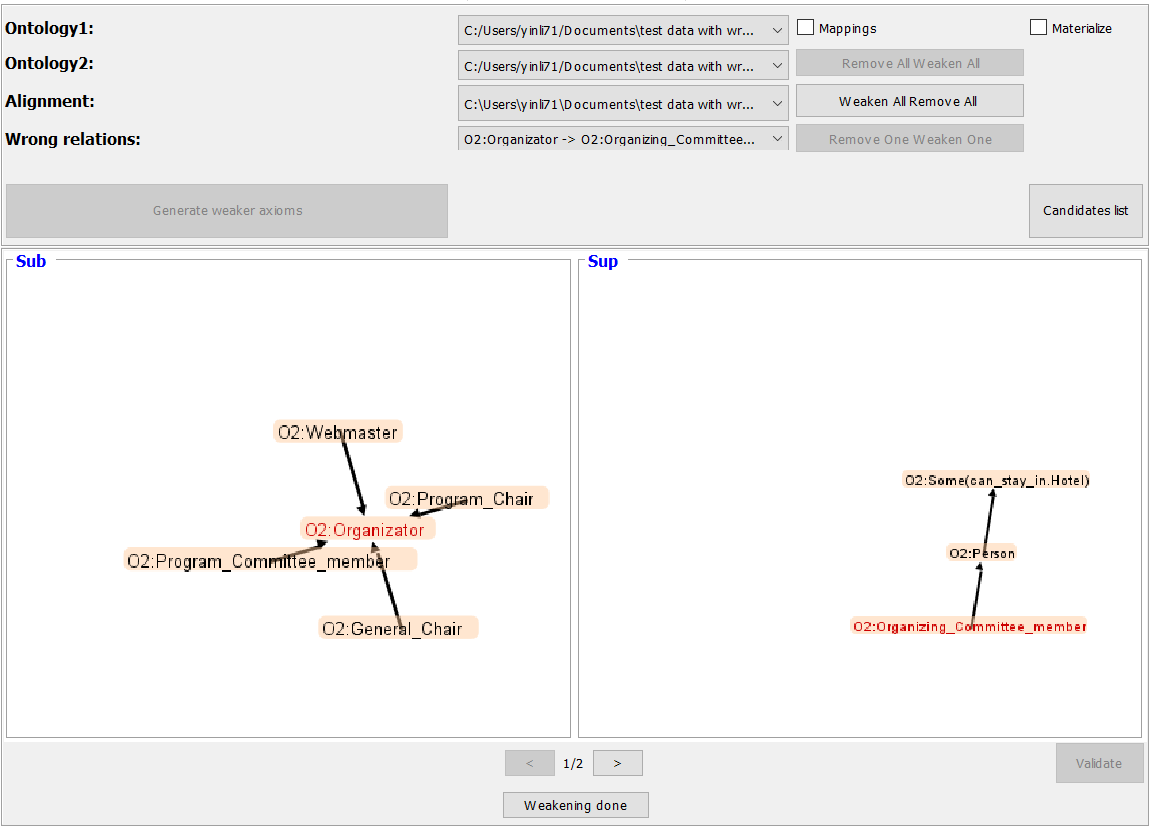}}\label{a}
\subfigure[]{\includegraphics[width=0.41\textwidth]{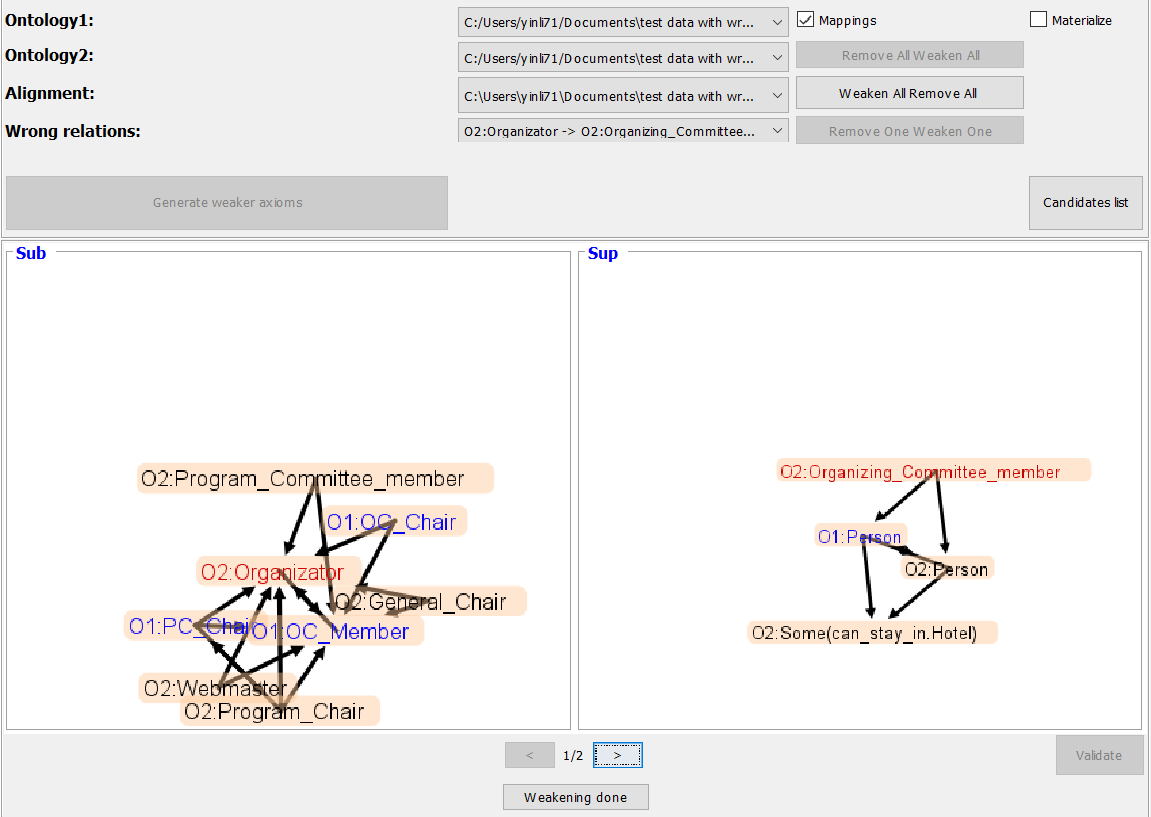}}\label{b}
\caption{Sub and sup when weakening wrong axiom Organizator $\sqsubseteq$ Organizing\_Committee\_Member in ontology network ekaw-sigkdd using (a) KB$_O$; (b) KB$_{ON}$. Concepts in wrong axiom in red, concepts in ekaw in blue, concepts in sigkdd in black. }
\label{UI-weakening}   
\end{center}
\vspace{-0.2cm}
\end{figure}

\begin{figure}[htb!] 
\begin{center}
 \vspace{-0.2cm}
\subfigure[]{\includegraphics[width=0.41\textwidth]{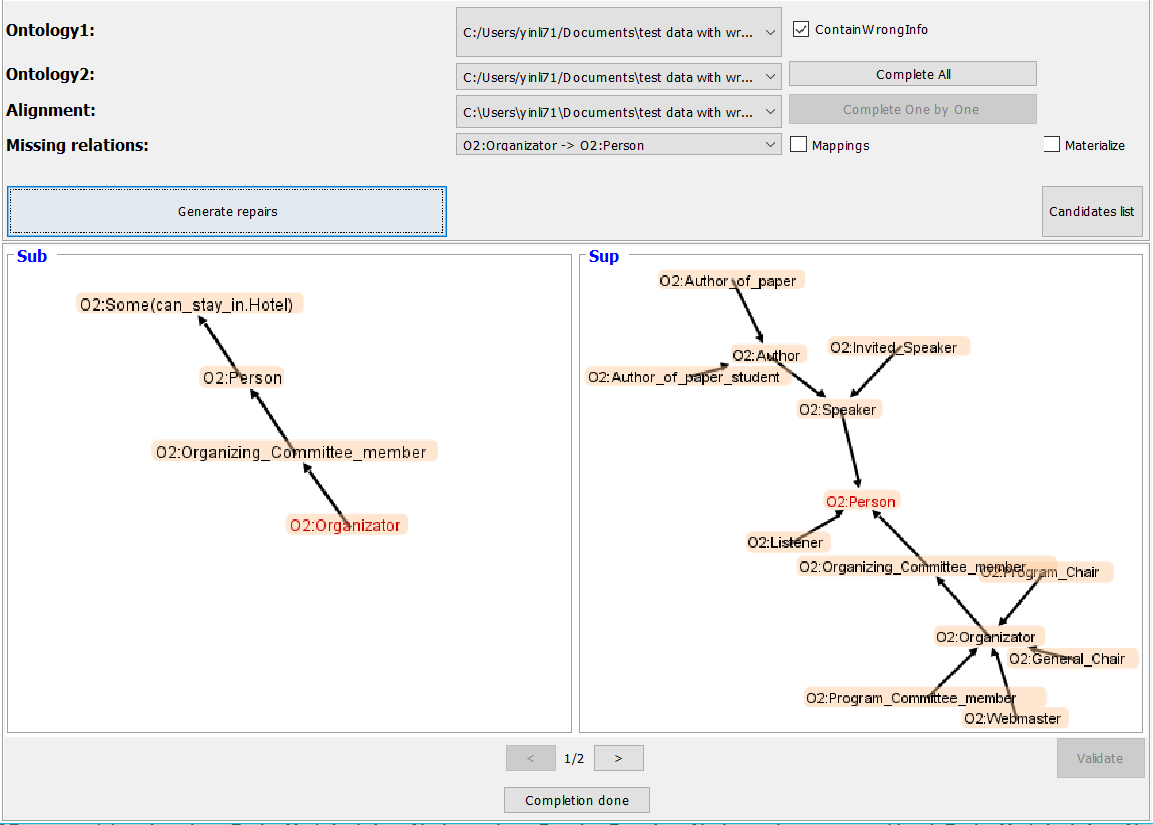}}\label{a}
\subfigure[]{\includegraphics[width=0.41\textwidth]{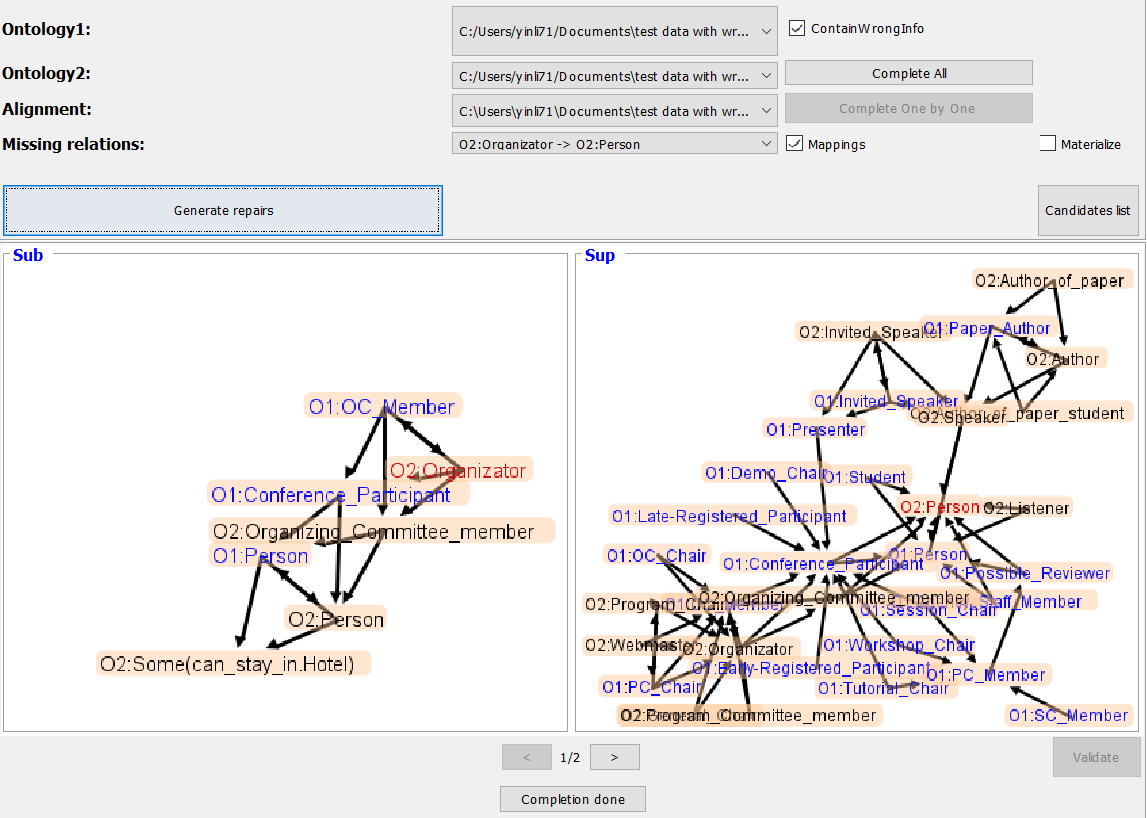}}\label{b}
\caption{Sub and sup when completing 
%wrong axiom Organizator $\sqsubseteq$ Organizing\_Committee\_Member 
weakened axiom Organizator $\sqsubseteq$ Person 
in ontology network ekaw-sigkdd using (a) KB$_O$; (b) KB$_{ON}$.
Concepts in weakened axiom in red, concepts in ekaw in blue, concepts in sigkdd in black. }
\label{UI-completion}   
\end{center}
\vspace{-0.2cm}
\end{figure}

For the basic operations the user interactions are adapted to the task at hand and different panels are used. For debugging the user requests the generation of the justifications. Then, the user can validate the axioms in the justifications or ask the system to compute Hitting sets and validate the axioms in those. The axioms to be validated are shown in a list.
During weakening, after choosing the preferred combination strategy, the user requests the system to generate the candidate weakened axioms for each axiom $\alpha$ $\sqsubseteq$ $\beta$ in the wrong asserted axioms set. The system computes the set of sub-concepts of $\alpha$ (sub) and the set of super-concepts of $\beta$ (sup), thereby representing the possible choices for weaker axioms. These weaker axioms can be visualized in two ways: (i) as a list of axioms and (ii) by two panes representing the sub and sup sets with their subsumption relations (Fig. \ref{UI-weakening}), respectively. In the latter case the user can choose weakened axioms by clicking on a concept in the sub set and a concept in the sup set and select the axiom as a weakened axiom. The advantage of this case is that these axioms are validated with the context in the domain of the ontology (showing the partial ontology through visualization). 
The set-up for completing is similar as for the weakening. For an axiom $\alpha$ $\sqsubseteq$ $\beta$ to be completed, the set of super-concepts of $\alpha$ and the set of sub-concepts of $\beta$ are computed and visualized as lists or using two panes (Fig. \ref{UI-completion}). Validation is performed in a similar way as for weakening.

%The system is available at %\url{https://figshare.com/s/f3b9472a7e5dd69237dc}.
%\url{https://figshare.com/s/e652a8e100579316f876}.

\section{Conclusion}

In this paper we presented a framework for repairing ontology networks that alleviates the problem of removing correct knowledge when removing unwanted axioms from the network. It uses the basic operators of debugging, removing, weakening and completing for which there exist different approaches. We defined combination operators that represent choices to be made when combining the basic operations. We introduced different levels of autonomy for the ontologies and alignments in the networks and defined combination operators based on these levels and the use of background knowledge during the computation of repairs and the selection of the final solution. We have shown the influence of the combination operators on the quality of the repaired network. The framework gives flexibility to the ontology and alignment owners regarding their use and update policies. The framework also provides a blueprint to extend existing systems to more general systems for repairing ontologies and ontology networks. 
 
\paragraph*{Supplemental Material Statement:}
The supplemental material is available at \url{https://figshare.com/s/e652a8e100579316f876}.  It includes the  ontology versions used in the experiments and implemented system.  

\section*{Acknowledgements}

This work is financially supported by the Swedish e-Science Research Centre (SeRC) and the EU Horizon Europe project Onto-DESIDE under Grant Agreement 101058682.

%\pagebreak

%\newpage

\bibliography{ref-short}
\bibliographystyle{splncs04}

\appendix

\vspace{0.5cm}

\noindent {\bf \Large Appendix}

 \vspace{0.5cm}
 
% This appendix is part of the supplemental material and is also available at \url{https://figshare.com/s/e652a8e100579316f876}.

\noindent  \vspace{0.5cm}
 
\section{Combinations of basic operations}

\subsection{Combination operators}

Table \ref{tab:operation-overview} shows combination operators that can be used as building blocks in the design of algorithms. For each of the basic operations (debugging, removing, weakening and completing) we show choices. The operations for removing, weakening and completing are taken from \cite{LL23}.
The operations for debugging are from \cite{LL24,LL24submitted}.

\begin{table}[htb]
\begin{center}
\caption{\label{tab:operation-overview}Debugging, removing, weakening and completing - operations.}
\begin{tabular}{|l|l|}
\hline
Operations & Description \\
\hline
S-one & Compute the justifications for one wrong axiom at the time. \\
S-all & Compute the justifications for all wrong axioms at once. \\
\hline
\hline
D-one-v & Generate one Hitting set from the justifications, then   validate \\ &the asserted axioms in the generated Hitting set.\\
D-v-one & Validate one valid Hitting set in the justifications.\\
D-all-v/D-v-all & Validate all asserted axioms from the justifications.\\
\hline
\hline
R-all & Remove all the wrong axioms at once.\\
R-one & Remove the wrong axioms one at a time.\\
R-none & Remove nothing.\\
\hline
\hline
W-all & Weaken all wrong axioms at once.\\
W-one & Weaken the wrong axioms one at a time\\
\hline
\hline
C-all & Complete all weakened axioms at once.\\
C-one & Complete the weakened axioms one at a time.\\
\hline
\hline
AB-one & Add one wrong axiom back.\\ 
AB-all & Add all wrong axioms back.\\
AB-none & Add nothing back.\\
\hline
\hline
U-now & Update the changes immediately. \\
U-end\_one & Update the changes after the iteration of each wrong axiom. \\
U-end\_all & Update the changes after iterations of all wrong axioms. \\
\hline
%\hline
%AS$_{ON}$&The axioms to be added to the network can be axioms within\\& the respective ontologies and  mappings between the ontologies.\\
%AS$_{O}$&The axioms to be added to the network can only be axioms\\& within the respective ontologies.\\
%AS$_{M}$ &The axioms to be added to the network can only be mappings\\& between the ontologies. \\
%\hline
\end{tabular}
\end{center}

\end{table}

\section{Derivation of Hasse diagrams}

The proofs for the Hasse diagrams are based on the following observations. 
For a given TBox $\mathcal{T}$, let Der($\mathcal{T}$) denote the set of derivable axioms from $\mathcal{T}$. Then, for TBoxes $\mathcal{T}_1$ and $\mathcal{T}_2$, if $\mathcal{T}_1$ $\subseteq$ $\mathcal{T}_2$, then we know that  Der($\mathcal{T}_1$) $\subseteq$ Der($\mathcal{T}_2$). This means that if an axiom is derivable from TBox $\mathcal{T}_1$, it is also derivable from TBox $\mathcal{T}_2$, but not necessarily the other way around.  Therefore, if $\mathcal{T}_1$ $\subseteq$ $\mathcal{T}_2$, all correct axioms in $\mathcal{T}_1$ can also be derived from $\mathcal{T}_2$, and thus $\mathcal{T}_2$ is more or equally complete than $\mathcal{T}_1$. Similarly, all wrong axioms in $\mathcal{T}_1$ can also be derived from $\mathcal{T}_2$, and thus $\mathcal{T}_2$ is more or equally incorrect than $\mathcal{T}_1$.  
\subsection{Ontologies}

The Hasse diagrams for removing, weakening and completing were derived in \cite{LL23}. The Hasse diagram for debugging was derived in \cite{LL24,LL24submitted}. The proofs for these are in \cite{LL24submitted} and are copied here for the sake of completeness.

\subsubsection{Debugging}

%Given a TBox $\mathcal{T}$ and two sets of wrong asserted axioms $D_1$ and $D_2$ such that $D_1 \sqsubseteq \mathcal{T}$, $D_2 \sqsubseteq \mathcal{T}$, and $D_1 \sqsubseteq D_2$, then $\mathcal{T}$ $\setminus$ $D_2$  $\sqsubseteq$ $\mathcal{T}$ $\setminus$ $D_1$. Given the reasoning above, the ontology where the wrong axioms in $D_2$ have been removed (represented by  $\mathcal{T}$ $\setminus$ $D_2$) is less or equally incorrect than the ontology where the wrong axioms in $D_1$ have been removed (represented by $\mathcal{T}$ $\setminus$ $D_1$). The latter is more or equally complete than the former.

When choosing the operations which contain D-all-v or D-v-all, all wrong asserted axioms in the justifications of the given wrong axioms are retained after validation %($W_{S-all,D-all-v}$ = $ W_{S-one,D-all-v}$ = $W_{S-all,D-v-all}$ = $W_{S-one,D-v-all}$)
($W_{S-all,D-all-v/D-v-all}$ = $ W_{S-one,D-all-v/D-v-all}$).\footnote{We consider here the debugging phase separate from the weakening and completing. In the case we would interleave the operations, it is not clear how to compare the incorrectness of the original ontology and the repaired ontology.
For instance, if we use S-one, then when removing the axioms in the justifications of a selected wrong axiom from the original ontology (version 1) we obtain a less incorrect ontology (version 2). Then during the weakening and completing steps, new axioms are added making the new version of the ontology (version 3) more complete, but possibly also more incorrect than version 2 (as new wrong axioms may be derivable using wrong axioms still in the ontology in combination with the added axioms). These new axioms may then influence the justifications for the next selected wrong axiom to process, finding more wrong asserted axioms to remove and thus find a less incorrect ontology (version 4) than version 3. Thus, when combining debugging with weakening and completing for S-one, for each original wrong axiom we would make the ontology less incorrect in one way and then more incorrect in another way, but also producing opportunities for additional removal of asserted wrong axioms that may not appear when using S-all.} For the other choices, not all axioms in the justifications are validated and used and thus the set of asserted wrong axioms to remove for each of those choices is a subset of $W_{S-all,D-all-v/D-v-all}$.
%$W_{S-all,D-all-v}$. 
Note that the Hitting sets computed by the different choices may be different and thus they are siblings in the Hasse diagram. 

\subsubsection{Removing} 

In general, when removing all axioms at once, the TBox is a subset of the TBox with one axiom removed, which in turn is a subset of the TBox where no axioms are removed. When adding no axioms back, the TBox is a subset of the TBox with one axiom added back, which in turn is a subset of the TBox where all axioms are added back.
If no wrong axioms are removed, then nothing needs to be added back and thus AB-one, AB-all and AB-none have the same result ($\mathcal{T}_{R-none,AB-all}$ = $\mathcal{T}_{R-none,AB-one}$ = $\mathcal{T}_{R-none,AB-none}$). The TBox for these strategies is larger during computation (of weakened or completed axiom sets) than the TBoxes where one or all wrong axioms are removed.
If one wrong axiom at the time is removed, the adding back all (AB-all) or one (AB-one) give the same result ($\mathcal{T}_{R-one,AB-all}$ = $\mathcal{T}_{R-one,AB-one}$) as both strategies add the same one axiom back. The TBox for these strategies is larger than when no wrong axiom is added back ($\mathcal{T}_{R-one,AB-none}$ $\subseteq$ $\mathcal{T}_{R-one,AB-all}$ = $\mathcal{T}_{R-one,AB-one}$).
When all wrong axioms are removed at once, then they will be added back at the end or not.\footnote{After completing they should be removed, but after weakening they could be added back for the completion step.} However, this does not influence the TBox during the computation. Therefore, the add back strategy does not matter and the TBox during computation is smaller than when wrong axioms were removed one at a time ($\mathcal{T}_{R-all,AB-all}$ = $\mathcal{T}_{R-all,AB-one}$ = $\mathcal{T}_{R-all,AB-none}$  $\subseteq$ $\mathcal{T}_{R-one,AB-none}$).

\subsubsection{Weakening} 

First, we note that updating immediately or updating after each wrong axiom is the same operation for weakening, as a complete weakened axiom set for a wrong axiom is computed. Thus, the TBox for ($\mathcal{T}_{W-one,U-now}$) is the same as for ($\mathcal{T}_{W-one,U-end\_one}$), and  the TBox for ($\mathcal{T}_{W-all,U-now}$) is the same as for ($\mathcal{T}_{W-all,U-end\_one}$).
Further, when weakening one axiom at a time and updating the TBox (i.e., adding the axioms of the weakened axiom set for a wrong axiom) immediately, results in a larger TBox for the next computations of weakened axiom sets for wrong axioms, than if we would not update immediately ($\mathcal{T}_{W-one,U-end\_all}$ $\subseteq$ $\mathcal{T}_{W-one,U-now}$). When not immediately updating, the TBox for generating the weakened axiom sets, stays the same for all wrong axioms and thus gives the same result as weakening all wrong axioms at once. Thus, $\mathcal{T}_{W-all,U-now}$ = $\mathcal{T}_{W-all,U-end\_all}$ = $\mathcal{T}_{W-one,U-end\_all}$.

\subsubsection{Completing} 

When completing one axiom at a time and updating the TBox (i.e., adding the axioms of the completed axiom set for a weakened axiom) immediately, results in a larger TBox for the next computations of completed axiom sets for weakened axioms than not updating immediately ($\mathcal{T}_{C-one,U-end\_one}$ $\subseteq$ $\mathcal{T}_{C-one,U-now}$, $\mathcal{T}_{C-one,U-end\_all}$ $\subseteq$ $\mathcal{T}_{C-one,U-now}$). When not updating immediately, there is the choice between updating after all weakened axioms for a particular wrong axiom have been processed or waiting until all weakened axioms for all wrong axioms are processed. The TBox for the former case is larger than the one for the latter case ($\mathcal{T}_{C-one,U-end\_all}$ $\subseteq$ $\mathcal{T}_{C-one,U-end\_one}$). Waiting to update the TBox until all weakened axioms for all wrong axioms are processed, means the TBox stays the same during the computation of the completed axioms sets and thus gives the same result as completing all weakened axioms at once ($\mathcal{T}_{C-one,U-end\_all}$ = $\mathcal{T}_{C-all,U-end\_all}$ = $\mathcal{T}_{C-all,U-end\_one}$ = $\mathcal{T}_{C-all,U-now}$).

\subsection{Ontology networks}

\subsubsection{Background knowledge}

Let ${\mathcal T}_1$, ...,${\mathcal T_n}$ be TBoxes representing ontologies ${\mathcal O_1}$, ..., ${\mathcal O_n}$, respectively. For $i, j \in [1..n]$ with $i < j$, let ${\mathcal A_{ij}}$ be an alignment between ontology ${\mathcal O_i}$ and ${\mathcal O_j}$. 
Let TBox $\mathcal{T}_{ON}$  =  ($\bigcup_{i=1..n}$ ${\mathcal T}_i$)  $\cup$ ($\bigcup_{i,j=1..n, i<j}$ ${\mathcal A_{ij}}$) represent the network. 

Let Tbox $\mathcal{T}_{MON}$ be the materialized version of the network, i.e., all (chosen) derived axioms in the network represented by $\mathcal{T}_{ON}$ are asserted in $\mathcal{T}_{MON}$.
Then $\mathcal{T}_{MON}$ =  ($\bigcup_{i=1..n}$ ${\mathcal T}_{mi}$)  $\cup$ ($\bigcup_{i,j=1..n, i<j}$ ${\mathcal A_{mij}}$) where ${\mathcal T}_{mi}$ contains all axioms in $\mathcal{T}_{MON}$ that relate concepts within ontology ${\mathcal O_i}$, and ${\mathcal A_{mij}}$ contains all mappings in $\mathcal{T}_{MON}$ between concepts in ${\mathcal O_i}$ and concepts in ${\mathcal O_j}$.

Then we have the following observations.\\
(i) Der($\mathcal{T}_{MON}$) = Der($\mathcal{T}_{ON}$).\\
(ii) As each axiom in ${\mathcal T}_{i}$ is derivable from the network, it is contained in ${\mathcal T}_{mi}$. Thus, for each i: ${\mathcal T}_{i}$ $\subseteq$ ${\mathcal T}_{mi}$.\\
(iii) As each mapping in ${\mathcal A_{ij}}$ is derivable from the network, it is contained in ${\mathcal A_{mij}}$. Thus, for each i,j: ${\mathcal A_{ij}}$ $\subseteq$ ${\mathcal A_{mij}}$.\\

We give now the proof of the Hasse diagrams considering that all ontologies (or alignments) use the same autonomy level. (The framework allows to have different autonomy levels for different ontologies and alignments. For this case the proofs are similar.)

If we use KB$_{O}$, then the algorithms for debugging, repairing and weakening use TBox $\bigcup_i$ ${\mathcal T}_{i}$. (No connections between the ontologies, so, in principle each ontology is dealt with separately.)
When using KB$_{MO}$, then the algorithms use TBox $\bigcup_i$ ${\mathcal Tm}_{i}$. 
By observation (ii), more knowledge is used for KB$_{MO}$ than for KB$_{O}$, and thus the repaired ontology network for KB$_{MO}$ is more (or equally) complete and more (or equally) 
 incorrect than for KB$_{O}$.

KB$_{ON}$ uses TBox $\mathcal{T}_{ON}$ which is logically equivalent to using $\mathcal{T}_{MON}$ (observation (i)). As $\bigcup_i$ ${\mathcal Tm}_{i}$ $\subseteq$ ($\bigcup_{i=1..n}$ ${\mathcal T}_{mi}$)  $\cup$ ($\bigcup_{i,j=1..n, i<j}$ ${\mathcal A_{mij}}$) = $\mathcal{T}_{MON}$, we know that the repaired ontology network for KB$_{ON}$ is more (or equally)  complete and more (or equally)  incorrect than for KB$_{MO}$.

Similar reasoning (using observations (i) and (iii)) leads to the fact that the repaired ontology network for KB$_{ON}$ is more (or equally) complete and more (or equally) incorrect than for KB$_{MM}$, which in its turn is more (or equally) complete and more (or equally) incorrect than for KB$_{M}$.

\subsubsection{Add sets}

Axioms are added to the ontology network after weakening, completing or as the final result.
Assume the set of axioms AS is a result of one of these stages. Then AS = OA $\cup$ M where OA is the set of axioms within ontologies in the result and M is the set of mappings between ontologies in the result. We know that OA $\cap$ M = $\emptyset$.

Assume the current ontology is represented by $\mathcal{T}$.
When using AS$_{ON}$, AS is added to the ontology network. When using AS$_{O}$, OA is added to the ontology network. Finally, when using AS$_{M}$, M is added to the ontology network.
As $\mathcal{T}$ $\cup$ M $\subseteq$ $\mathcal{T}$ $\cup$ AS and $\mathcal{T}$ $\cup$ OA $\subseteq$ $\mathcal{T}$ $\cup$ AS, using  AS$_{ON}$ leads to a more complete network than using AS$_{O}$ or AS$_{M}$.

\section{Computing weakened and completed axioms}

In this section we show the sub- and super-concept sets used for computing weakened and completed axioms for the example in Sect. \ref{sec-combinations}. (When there are empty cells in the tables, it means that we did not need to compute the sub- and super-concepts sets for these choices of combination operator.)

\begin{table}[H]
\begin{center}
\vspace{-0.2cm} 
%\begin{tabular}{|l|l|l|}
\caption{The sub- and super-concepts computed during weakening e $\sqsubseteq$ b.}
\begin{tabular}{|p{1.8cm}|l|l|l|}
\hline
  & KB$_O$ &KB$_{MO}$ & KB$_{ON}$ \\\hline
Sub($e$,${\mathcal T}$)&\{e, f\}&\{e, f\}&\{e, f, E, F\} \\
Sup($b$,${\mathcal T}$)&\{b\}&\{b, a\}&\{b, a, D, B, A\}\\\hline
\end{tabular}
\vspace{-0.5cm} 
\label{example-result4} 
\end{center}
\end{table}

\begin{table}[H]
\begin{center}
\vspace{-0.2cm} 
%\begin{tabular}{|l|l|l|}
\caption{The sub- and super-concepts computed during weakening b $\sqsubseteq$ D.}
\begin{tabular}{|p{1.8cm}|l|l|l|}
\hline
  & KB$_M$ &KB$_{MM}$ & KB$_{ON}$ \\\hline
Sub($b$,${\mathcal T}$)&\{b, D\}&\{e, f, b, D, E, F\}  &\{e, f, b, D, E, F\}  \\
Sup($D$,${\mathcal T}$)&\{D, b\}&\{D, b, A, a, B\} &\{D, A, B, b, a\} \\\hline
\end{tabular}
\vspace{-0.5cm} 
\label{example-result3} 
\end{center}
\end{table}

\begin{table}[H]
\begin{center}
\vspace{-0.2cm} 
%\begin{tabular}{|l|l|l|}
\caption{The sub- and super-concepts computed during completing the weakened axioms for wrong axiom e $\sqsubseteq$ b.}
\begin{tabular}{|p{1.8cm}|p{0.8cm}|l |l|}
\hline
Background knowledge base & KB$_O$ &KB$_{MO}$ & KB$_{ON}$ \\\hline
Sup($f$,${\mathcal T}$)&\{e, f\}&\{e, f, b, a\} &\{A, B, C, D, E, F, e, f, b, a\}\\
Sub($b$,${\mathcal T}$)&\{b\} &\{e, f, b\}&\{D, E, F, e, f, b\}  \\\hline
Sup($e$,${\mathcal T}$)&&\{e, b, a\} &\{e, E, b, D, A, B, a\}\\
Sub($a$,${\mathcal T}$)&&\{d, e, f, b, a\} &\{b, C, f, d, e, E, F, D, A, B, a\} \\\hline
Sup($E$,${\mathcal T}$)& & &\{e, E, b, D, A, B, a\}\\
Sub($A$,${\mathcal T}$)& & &\{b, C, f, d, e, E, F, D, A, B, a\} 
\\\hline
\end{tabular}
\vspace{-0.5cm} 
\label{example-result5} 
\end{center}
\end{table}

\begin{table}[H]
\begin{center}
\vspace{-0.2cm} 
%\begin{tabular}{|l|l|l|}
\caption{The sub- and super-concepts computed during completing the weakened axioms for wrong mapping b $\sqsubseteq$ D.}
\begin{tabular}{|p{1.8cm}|l|l |l|}
\hline
Background knowledge base & KB$_M$ &KB$_{MM}$ & KB$_{ON}$ \\\hline
Sup($b$,${\mathcal T}$)& &\{b, a, B, A, D\} &\{A, B, D  b, a\}\\
Sub($B$,${\mathcal T}$)&  &\{B, D, E, F, e, f, b\}&\{B, D, E, F, e, f, b\}  \\\hline
Sup($e$,${\mathcal T}$)&&  &\{e, E, b, D, A, B, a\}\\
Sub($a$,${\mathcal T}$)&& &\{b, C, f, d, e, E, F, D, A, B, a\} \\\hline
\end{tabular}
\vspace{-0.5cm} 
\label{example-result6} 
\end{center}
\end{table}

\section{Experiments - results}

In the experiments we use 5 ontology networks. Each network consists of two ontologies and an alignment. The number of concepts, roles and axioms of the ontologies are given in Table \ref{tab:ontologies} and the number of mappings in the alignments is given in Table \ref{tab:mappings}.
Tables \ref{tab:wrong axioms} and \ref{tab:wrong mappings} list the wrong axioms and mappings we introduced in each ontology network. These wrong axioms were generated by replacing existing axioms
with axioms where their left/right-hand side concepts were changed.

The results of the experiments are listed in Tables \ref{ekaw-sigkdd}-\ref{iasted-sigkdd}. These tables list the sizes of the sub-concepts set of $\alpha$ and the super-concepts set of $\beta$ for each wrong axiom $\alpha \sqsubseteq \beta$ and the correct weakened axioms. Further, for each correct weakened axiom  $\alpha \sqsubseteq \beta$, it shows the sizes of the super-concepts set of $\alpha$ and the sub-concepts set of $\beta$ as well as the correct completed axioms. The sizes of these sets reflect the number of axioms that need to be validated by the oracle. We note, however, that using the visualization in our system these sets are shown together and thus the validation of many axioms can be done at the same time.

 \begin{table}[htb]
\begin{center}
\caption{\label{tab:ontologies}Ontologies}
%\begin{tabular}{|p{1.05cm}|p{1.5cm}|p{0.7cm}|p{0.9cm}|p{1.1cm}|p{0.8cm}|p{0.8cm}|}
\begin{tabular}{|l|r|r|r|r|}
\hline
&ekaw&sigkdd&cmt&iasted\\\hline
Concepts&74&49&36&140\\\hline
Roles&33&17&49&38\\\hline 
Axioms&340&193&319&539\\\hline
\end{tabular}
\end{center}
\end{table}

\begin{table}[htb]
\begin{center}
\caption{\label{tab:mappings}Ontology networks}
%\begin{tabular}{|p{1.05cm}|p{1.5cm}|p{0.7cm}|p{0.9cm}|p{1.1cm}|p{0.8cm}|p{0.8cm}|}
\begin{tabular}{|l|r|r|r|r|r|}
\hline
Ontology network&ekaw-sigkdd&cmt-sigkdd&cmt-ekaw&iasted-sigkdd&ekaw-iasted\\\hline
Mappings&11&9&11&14&10\\\hline
\end{tabular}
\end{center}
\end{table}

\begin{table}[H]
\begin{center}
\vspace{-0.2cm}     
%\begin{tabular}{|l|p{3.2cm}|p{3.2cm}|p{3.2cm}|p{3.2cm}|}
\caption{\label{tab:wrong axioms}Wrong axioms in each ontology.}
\begin{tabular}{|l|p{8.5cm}|}
\hline
Ontology&Wrong axioms ($W$) \\\hline
ekaw& Tutorial $\sqsubseteq$ Conference\\\hline
sigkdd&Organizator $\sqsubseteq$  Organizing\_Committee\_Member\\\hline
cmt&Program \_Committee $\sqsubseteq$ Person

Program\_ Committee\_Chair $\sqsubseteq$ Program\_ Committee\\\hline
iasted& Final\_manuscript $\sqsubseteq$ Publication\\\hline
\end{tabular}
\vspace{-0.2cm} 
\end{center}
\end{table}

\begin{table}[H]
\begin{center}
%\vspace{-0.2cm}     
%\begin{tabular}{|l|p{3.2cm}|p{3.2cm}|p{3.2cm}|p{3.2cm}|}
\caption{\label{tab:wrong mappings}Wrong mappings in each ontology network.}
\begin{tabular}{|p{2.9cm}|p{9.2cm}|}
\hline
Ontology network&Wrong mappings ($W$) \\\hline
ekaw-sigkdd& sigkdd:Organizator $\sqsubseteq$ ekaw:OC\_Member\\\hline
cmt-sigkdd&cmt:Conference\_Chair $\sqsubseteq$ sigkdd:Program\_Chair\\\hline
cmt-ekaw&cmt:Conference Member $\sqsubseteq$ ekaw:OC\_Member\\\hline
ekaw-iasted&ekaw:Workshop\_ Chair $\sqsubseteq$ iasted:Session\_Chair\\\hline
iasted-sigkdd&iasted:Registration\_fee $\sqsubseteq$ sigkdd:NonMember\_Registration\\\hline
\end{tabular}
\vspace{-0.5cm} 
\end{center}
\end{table}

\begin{table}[H]
\begin{center}
\vspace{-0.2cm} 
%\begin{tabular}{|l|l|l|}
\caption{ekaw-sigkdd}
\begin{tabular}{|p{1.4cm}|p{2.05cm}|p{2.05
cm} |p{1.7 cm}|p{0.8 cm}|p{0.7cm}|p{1.9cm} |p{3cm}|}
\hline
 %& (OC\_Member, Organizator)               & Organizator $\sqsubseteq$ Organizing\_Committee\_Member            \\\hline
Wrong  & \multicolumn{2}{ p{4.1cm}|}{sigkdd:Organizator$\sqsubseteq$
sigkdd:Organizing\_Committee \_Member}
&\multicolumn{2}{ p{2 cm}|}{ ekaw:Tutorial$\sqsubseteq$ ekaw:Conference} & \multicolumn{3}{ p{4cm}|}{sigkdd:Organizator$\sqsubseteq$ekaw:OC\_Member} \\\hline
KB&KB$_{O}$/KB$_{MO}$&KB$_{ON}$&KB$_{O}$/KB$_{MO}$&KB$_{ON}$&KB$_{M}$&KB$_{MM}$&KB$_{ON}$ \\\hline
$\mid$Sub($\alpha$,${\mathcal T}$)$\mid$&5& 8  & 1& 1 & 2& 4& 8  \\
$\mid$Sup($\beta$,${\mathcal T}$)$\mid$&3& 4 &3& 4 &2&4&7\\\hline
Weakened & \multicolumn{2}{ p{4cm}|}{sigkdd:Organizator $\sqsubseteq $sigkdd:Person}&\multicolumn{2}{ p{2 cm}|}{ekaw:Tutorial $\sqsubseteq $ekaw:Scientific \_Event } &&sigkdd:Organ izator$\sqsubseteq$ ekaw:Person& sigkdd:Organizator$\sqsubseteq$ ekaw:Conference \_Participant \\\hline
$\mid$Sup($\alpha$,${\mathcal T}$)$\mid$&4&7  &4&5  & &4&7  \\
$\mid$Sub($\beta$,${\mathcal T}$)$\mid$&13& 32 &15& 16 &&10&18  \\\hline
completed &sigkdd: Organizator$\sqsubseteq $ sigkdd:Person,
sigkdd: Organizing \_Committee \_member$\sqsubseteq$ sigkdd: Organizator &sigkdd: Organizator$\sqsubseteq$ sigkdd:Person, sigkdd: Organizing \_Committee \_member$\sqsubseteq$ sigkdd: Organizator, sigkdd: Organizing\_ Committee \_Member$\sqsubseteq$ ekaw:OC \_Member&\multicolumn{2}{ p{2 cm}|}{ ekaw:Tutorial$\sqsubseteq$ ekaw:Individual \_Presentation}&&sigkdd:Organ izator$\sqsubseteq$ ekaw:Person&ekaw:OC\_Member$\sqsubseteq$ sigkdd:Organizing
\_Committee\_Member, sigkdd:Organizator$\sqsubseteq$ ekaw:Conference \_Participant\\\hline

\end{tabular}
\vspace{-0.5cm} 
\label{ekaw-sigkdd} 
\end{center}
\end{table}

\begin{table}[H]
\begin{center}
%\begin{tabular}{|p{2.4cm}|p{3.5cm}|p{3.5cm} |}
\vspace{-0.2cm} 
\caption{cmt-sigkdd}
\begin{tabular}{|p{1.4cm}|p{0.55cm}|p{0.85cm}|p{0.8cm}|p{0.8cm}|p{0.75cm}|p{0.6cm}|p{1.85cm}|p{2.15cm}|p{1.75cm}|p{1.75cm}|}
\hline
Wrong   &\multicolumn{3}{ p{2.2cm}|}{sigkdd:Organizator $\sqsubseteq$ sigkdd:Organizing \_Committee \_Member}&\multicolumn{2}{ p{1.5cm}|}{cmt:Program \_Committee$\sqsubseteq$ cmt:Person }&\multicolumn{3}{ p{3.9cm}|}{cmt:Conference\_Chair$\sqsubseteq$ 
sigkdd:Program\_Chair}&\multicolumn{2}{ p{2.8cm}|}{cmt:Program\_Committee \_Chair$\sqsubseteq$ cmt:Program\_Committee} \\\hline
KB&KB$_{O}$&KB$_{MO}$&KB$_{ON}$&KB$_{O}$/ KB$_{MO}$&KB$_{ON}$&KB$_{M}$&KB$_{MM}$&KB$_{ON}$&KB$_{O}$/ KB$_{MO}$&KB$_{ON}$ \\\hline
$\mid$Sub($\alpha$,${\mathcal T}$)$\mid$ & \multicolumn{2}{ p{1.3cm}|}{5}&7  &2&3&2&2& 2  &1& 1 \\
$\mid$Sup($\beta$,${\mathcal T}$)$\mid$&\multicolumn{2}{ p{1.3cm}|}{3}& 4   &1& 3 &2&4&8  &2&6\\\hline
Weakened & \multicolumn{3}{ p{2.2cm}|}{sigkdd:Organizator $\sqsubseteq$ sigkdd:Person}&\multicolumn{2}{ p{1.5cm}|}{}&&cmt:Conferen ce\_Chair$\sqsubseteq$ sigkdd:Person&cmt:Conference
\_Chair$\sqsubseteq$ sigkdd:Organiza tor& \multicolumn{2}{ p{2.6cm}|}{cmt:Program\_Committee \_Chair$\sqsubseteq$ cmt:Person} \\\hline
$\mid$Sup($\alpha$,${\mathcal T}$)$\mid$&4&4&5&\multicolumn{2}{ p{1.5cm}|}{}&&4&8&3&7 \\
$\mid$Sub($\beta$,${\mathcal T}$)$\mid$&13&14& 30&\multicolumn{2}{ p{1.5cm}|}{}& &10&7&16&30  \\\hline
completed & \multicolumn{3}{ p{2.2cm}|}{sigkdd:Organizing\_
Committee \_Member$\sqsubseteq$ sigkdd:Organizator,  sigkdd:Organizator $\sqsubseteq$ sigkdd:Person}&\multicolumn{2}{ p{1.5cm}|}{}&&cmt:Conferen ce\_Chair$\sqsubseteq$ sigkdd:Person&cmt:Conference \_Chair$\sqsubseteq$ sigkdd:General \_Chair&cmt:Program
\_Committee \_Chair $\sqsubseteq$ cmt:Program
\_Committee \_Member&cmt:Program
\_Committee \_Chair$\sqsubseteq$ cmt:Program
\_Committee \_Member, cmt:Program
\_Committee
\_Chair$\sqsubseteq$ sigkdd: Program
\_Chair\\\hline

\end{tabular}
\vspace{-0.5cm} 
\label{cmt-sigkdd}
\end{center}
\end{table}

\begin{table}[H]
\begin{center}
\vspace{-0.2cm} 
\caption{cmt-ekaw}
%\begin{tabular}{|p{2.4cm}|p{3.5cm}|p{3.5cm} |}
\begin{tabular}{|p{1.4cm}|p{1 cm}|p{1 cm}|p{1cm}|p{0.9cm}|p{0.7cm}|p{2.1cm}|p{2.3cm}|p{1cm}|p{1.cm}|}
\hline
Wrong  &\multicolumn{2}{ p{2.5cm}|}{cmt:Program \_Committee\_Chair $\sqsubseteq$ cmt:Program \_Committee}&\multicolumn{2}{ p{1.5cm}|}{cmt:Program\_ Committee$\sqsubseteq$ cmt:Person} & \multicolumn{3}{ p{2.6cm}|}{cmt:Conference\_Member$\sqsubseteq$ ekaw:OC\_Member} & \multicolumn{2}{ p{2.2cm}|}{ekaw:Tutorial$\sqsubseteq$ ekaw:Conference} \\\hline
KB&KB$_{O}$/ KB$_{MO}$&KB$_{ON}$&KB$_{O}$/ KB$_{MO}$&KB$_{ON}$&KB$_{M}$&KB$_{MM}$&KB$_{ON}$&KB$_{O}$/ KB$_{MO}$&KB$_{ON}$ \\\hline
$\mid$Sub($\alpha$,${\mathcal T}$)$\mid$ & 1&1  &2&2&2&4&12 &1&1\\
$\mid$Sup($\beta$,${\mathcal T}$)$\mid$&2&3  &1&2&2&4&5 &3&7\\\hline
Weakened &\multicolumn{2}{ p{2.4cm}|}{cmt:Program \_Committee\_Chair $\sqsubseteq$ cmt:Person}&\multicolumn{2}{ p{1.5cm}|}{}&&cmt:Conference \_Member$\sqsubseteq$ ekaw:Person&cmt:Conference \_Member$\sqsubseteq$ ekaw:Conference \_Participant&\multicolumn{2}{ p{2cm}|}{ ekaw:Tutorial$\sqsubseteq$ ekaw:Scientific \_Event} \\\hline
%Sup($\alpha$,${\mathcal T}$)&3(4)&&4 4 (5)&4(8) \\
%Sub($\beta$,${\mathcal T}$)&16(35)&&21 22 (22)&15(16)  \\\hline
$\mid$Sup($\alpha$,${\mathcal T}$)$\mid$&3&4&\multicolumn{2}{ p{1.5cm}|}{}&&4 &5&4&8 \\
$\mid$Sub($\beta$,${\mathcal T}$)$\mid$&16&35&\multicolumn{2}{ p{1.5cm}|}{}& &6& 22&15&16  \\\hline

completed &\multicolumn{2}{ p{2.5cm}|}{cmt:Program \_Committee\_Chair $\sqsubseteq$ cmt:Program \_Committee\_Chair}&\multicolumn{2}{ p{1.5cm}|}{}&&cmt:Conference \_Member$\sqsubseteq$ ekaw:Person&cmt:Conference \_Member$\sqsubseteq$ ekaw:Conference \_Participant, ekaw:Conference \_Participant$\sqsubseteq$ cmt:Conference \_Member&\multicolumn{2}{ p{2cm}|}{ekaw:Tutorial$\sqsubseteq$ ekaw:Individual\_ Presentation}\\\hline

\end{tabular}
\vspace{-0.5cm} 
\label{cmt-ekaw}
\end{center}
\end{table}

\begin{table}[H]
\begin{center}
\vspace{-0.2cm} 
\caption{ekaw-iasted}
%\begin{tabular}{|l|l|l|}
\begin{tabular}{|p{1.4cm}|p{1 cm}|p{0.93 cm}|p{0.65cm}|p{1.5cm}|p{2.1cm}|p{1cm}|p{0.8cm}|p{0.65cm}|p{1.7cm}|p{1.7cm}|}
\hline
 %& (OC\_Member, Organizator)               & Organizator $\sqsubseteq$ Organizing\_Committee\_Member            \\\hline
Wrong   &\multicolumn{2}{ p{1.92cm}|}{ekaw:Tutorial$\sqsubseteq$ ekaw:Conference }&\multicolumn{3}{ p{2.6cm}|}{ekaw:Workshop\_Chair$\sqsubseteq$ iasted:Session\_Chair}&\multicolumn{2}{ p{1.8cm}|}{iasted:Final\_ manuscript$\sqsubseteq$ iasted: Publication}&\multicolumn{3}{ p{2.2cm}|}{iasted:Session\_Chair$\sqsubseteq$ ekaw:Workshop\_Chair}\\\hline
KB&KB$_{O}$/ KB$_{MO}$&KB$_{ON}$&KB$_{M}$&KB$_{MM}$&KB$_{ON}$&KB$_{O}$/ KB$_{MO}$&KB$_{ON}$&KB$_{M}$&KB$_{MM}$&KB$_{ON}$ \\\hline
$\mid$Sub($\alpha$,${\mathcal T}$)$\mid$&1&2 &  2&2&2 & 1&1& 2&2&2 \\
$\mid$Sup($\beta$,${\mathcal T}$)$\mid$&3&4& 2&4&16  &3&3& 2&4&16\\\hline
Weakened &\multicolumn{2}{ p{1.92cm}|}{ekaw:Tutorial$\sqsubseteq$ ekaw:Scientific \_Event}&&ekaw:Work shop\_Chair $\sqsubseteq$ iasted:Pers on&ekaw:Workshop \_Chair$\sqsubseteq$ iasted:Delegate&\multicolumn{2}{ p{1.8cm}|}{iasted:Final\_ manuscript$\sqsubseteq$ iasted:Item}&&iasted: Session
\_Chair$\sqsubseteq$ ekaw:Person&iasted: Session
\_Chair$\sqsubseteq$ ekaw: Conference \_Participant   \\\hline
%Sup($\alpha$,${\mathcal T}$)  &4(12) &14(16) &5(6)&3 13 (16)  \\
%Sub($\beta$,${\mathcal T}$)  &15(18)&23(24) &34(61)&13 13 (13)  \\\hline
$\mid$Sup($\alpha$,${\mathcal T}$)$\mid$  &4&12 &&4& 16 &5&6& &4&16  \\
$\mid$Sub($\beta$,${\mathcal T}$)$\mid$  &15&18& &6&24 &34&61& &6&13  \\\hline
completed &\multicolumn{2}{ p{1.92cm}|}{ekaw:Tutorial$\sqsubseteq$ ekaw:Individual \_Presentation}&&ekaw:Work shop\_Chair $\sqsubseteq$ iasted: Person& ekaw:Workshop \_Chair$\sqsubseteq$ iasted:Iasted\_ member&\multicolumn{2}{ p{1.8cm}|}{iasted:Final\_ manuscript$\sqsubseteq$ iasted: Submission}&&iasted: Session
\_Chair$\sqsubseteq$ ekaw:Person&iasted: Session \_Chair$\sqsubseteq$ ekaw:Session \_Chair\\\hline

\end{tabular}

\label{ekaw-iasted}
\vspace{-0.5cm} 
\end{center}
\end{table}

\begin{table}[H]
\begin{center}
\vspace{-0.2cm} 
%\begin{tabular}{|l|l|l|}
\caption{iasted-sigkdd}
\begin{tabular}{|p{1.4cm}|p{1.1cm}|p{1.1 cm} |p{0,65cm}|p{1.8cm}|p{2.65cm}|p{1.4cm} |p{1.4cm}|}
\hline
 %& (OC\_Member, Organizator)               & Organizator $\sqsubseteq$ Organizing\_Committee\_Member            \\\hline
Wrong   & \multicolumn{2}{ p{2.1cm}|}{iasted:Final \_manuscript$\sqsubseteq$ iasted:Publication }&\multicolumn{3}{ p{2.8cm}|}{iasted:Registration\_fee$\sqsubseteq$ sigkdd:Registration\_NonMember}& \multicolumn{2}{ p{3.1cm}|}{sigkdd:Organizator$\sqsubseteq$ sigkdd:Organizing
\_Committee\_Member} \\\hline
KB&KB$_{O}$/ KB$_{MO}$&KB$_{ON}$&KB$_{M}$&KB$_{MM}$&KB$_{ON}$&KB$_{O}$/ KB$_{MO}$&KB$_{ON}$ \\\hline
$\mid$Sub($\alpha$,${\mathcal T}$)$\mid$&1&1 & 2 &4&9& 5&5 \\
$\mid$Sup($\beta$,${\mathcal T}$)$\mid$&3&3&2 &4&9&3&4\\\hline

Weakened &\multicolumn{2}{ p{2.1cm}|}{iasted:Final \_manuscript$\sqsubseteq$ iasted:Item }&&iasted:Regis tration\_fee$\sqsubseteq$ sigkdd:fee &iasted:Registra tion\_fee$\sqsubseteq$ sigkdd:Registra tion\_fee, iasted:Nonmem ber\_Registration \_fee$\sqsubseteq$ sigkdd:Registra tion\_fee\_Non \_Member & \multicolumn{2}{ p{2.7cm}|}{sigkdd:Organizator$\sqsubseteq$ sigkdd:Person} \\\hline
%Sup($\alpha$,${\mathcal T}$)&5(5)  &12 12 (12)  8(9)  &4(5)  \\
%Sub($\beta$,${\mathcal T}$)&34(35)  &2 3 (9)   12(12) &13(44)  \\\hline
$\mid$Sup($\alpha$,${\mathcal T}$)$\mid$&5&5  & &4&12  9  &4 &5  \\
$\mid$Sub($\beta$,${\mathcal T}$)$\mid$&34&35 & &6& 9  12&13 &44  \\\hline
completed &\multicolumn{2}{ p{2.1cm}|}{iasted:Final \_manuscript$\sqsubseteq$ iasted:Submission}&&iasted:Regis tration\_fee$\sqsubseteq$ sigkdd:fee &iasted:Nonmember
\_Registration\_fee 
$=$
sigkdd:Registration \_fee\_Non\_Member, sigkdd:Registration \_fee $=$ iasted:Registration \_fee&\multicolumn{2}{ p{2.3cm}|}{sigkdd:Organizator$\sqsubseteq$ sigkdd:Person, sigkdd:Organizing
\_Committee\_member$\sqsubseteq$ sigkdd:Organizator} \\\hline

\end{tabular}

\label{iasted-sigkdd}
\vspace{-0.5cm} 
\end{center}
\end{table}

\end{document}